\documentclass{article}

\usepackage{microtype}
\usepackage{graphicx}
\usepackage{booktabs} 

\usepackage{hyperref}

\usepackage[accepted]{icml2023}

\usepackage{amsmath}
\usepackage{amssymb}
\usepackage{mathtools}
\usepackage{amsthm}
\usepackage{multirow}
\usepackage{latexsym}
\usepackage{graphicx}
\usepackage{epstopdf}
\usepackage{mathrsfs}
\usepackage{amsmath}
\usepackage{amssymb}
\usepackage{bm}
\usepackage{multirow}
\usepackage{booktabs}
\usepackage{algorithm}
\usepackage{algorithmic}
\usepackage{appendix}
\usepackage{verbatim}
\usepackage{caption}
\usepackage{subcaption}
\usepackage{xspace}
\usepackage{inconsolata}
\usepackage{makecell}

\usepackage[capitalize,noabbrev]{cleveref}

\theoremstyle{plain}

\theoremstyle{definition}

\theoremstyle{remark}

\usepackage[textsize=tiny]{todonotes}

\usepackage{array}
\newcommand{\PreserveBackslash}[1]{\let\temp=\\#1\let\\=\temp}
\newcolumntype{C}[1]{>{\PreserveBackslash\centering}p{#1}}
\newcolumntype{R}[1]{>{\PreserveBackslash\raggedleft}p{#1}}
\newcolumntype{L}[1]{>{\PreserveBackslash\raggedright}p{#1}}

\definecolor{red}{RGB}{255, 189, 189}
\definecolor{lue}{RGB}{135, 206, 250}
\definecolor{green}{RGB}{205, 255, 204}

\newcommand{\eg}{e.g.,\xspace}
\newcommand{\ie}{i.e.,\xspace}

\newcommand{\modelname}{\emph{Iterative Controlled Extrapolation}\xspace}
\newcommand{\modelacc}{\emph{ICE}\xspace}

\newcommand{\suptrain}{\textrm{sup-train}}
\newcommand{\unsup}{\textrm{unsup}}

\usepackage{amsmath,amsfonts,bm}

\def\eqref#1{equation~\ref{#1}}

\def\1{\bm{1}}

\def\vx{{\bm{x}}}

\DeclareMathAlphabet{\mathsfit}{\encodingdefault}{\sfdefault}{m}{sl}
\SetMathAlphabet{\mathsfit}{bold}{\encodingdefault}{\sfdefault}{bx}{n}

\def\gD{{\mathcal{D}}}

\icmltitlerunning{Extrapolation in Controlled Sequence Generation via Iterative Refinement} 

\begin{document}

\twocolumn[
\icmltitle{Extrapolative Controlled Sequence Generation via Iterative Refinement}



\icmlsetsymbol{equal}{*}

\begin{icmlauthorlist}
\icmlauthor{Vishakh Padmakumar}{nyu}
\icmlauthor{Richard Yuanzhe Pang}{nyu}
\icmlauthor{He He}{nyu}
\icmlauthor{Ankur P. Parikh}{google}
\end{icmlauthorlist}

\icmlaffiliation{nyu}{New York University}
\icmlaffiliation{google}{Google DeepMind}

\icmlcorrespondingauthor{Vishakh Padmakumar}{vishakh@nyu.edu}

\icmlkeywords{Machine Learning, ICML}

\vskip 0.3in

]  



\printAffiliationsAndNotice{}  

\begin{abstract}
We study the problem of extrapolative controlled generation, i.e., generating sequences with attribute values beyond the range seen in training. This task is of significant importance in automated design, especially drug discovery, where the goal is to design novel proteins that are \textit{better} (e.g., more stable) than existing sequences.
Thus, by definition, the target sequences and their attribute values are out of the training distribution,
posing challenges to existing methods that aim to directly generate the target sequence. Instead, in this work, we propose \modelname (\modelacc) which iteratively makes local edits to a sequence to enable extrapolation.
Specifically, we train the model on synthetically generated sequence pairs that demonstrate small improvement in the attribute value. Results on one natural language task (sentiment analysis) and two protein engineering tasks (ACE2 stability and AAV fitness) show that \modelacc considerably outperforms state-of-the-art approaches despite its simplicity.\footnote{Our code and models are available at \url{https://github.com/vishakhpk/iter-extrapolation}.}

\end{abstract}

\section{Introduction}
\label{sec:intro}

\begin{figure*}[t!]
    \centering
    \includegraphics[width=0.8\textwidth]{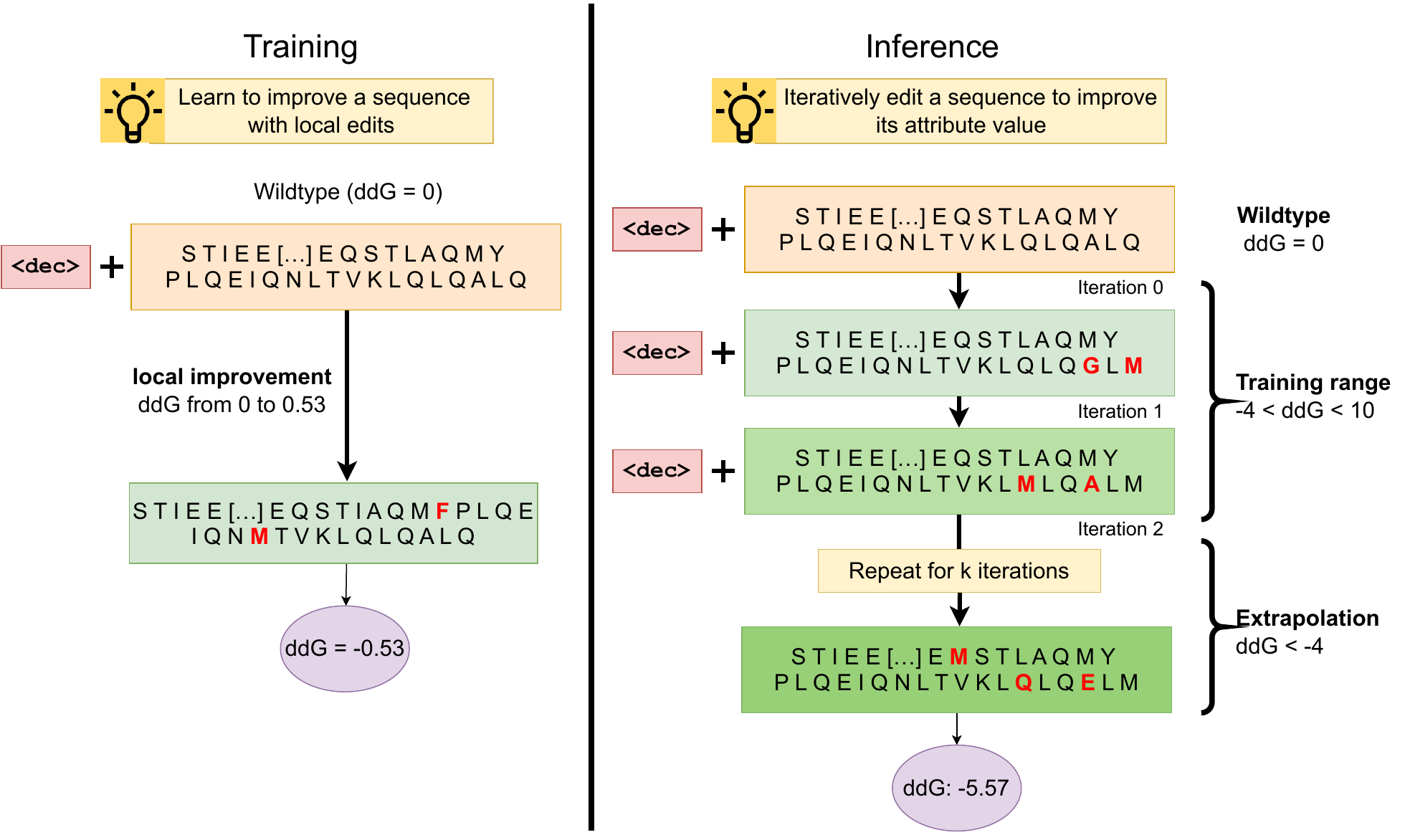}
    \caption{An overview of the approach, \modelname (\modelacc), on the ACE2 stability task. Our initial dataset only contains proteins with {ddG} values (lower means more stable) 
    between -4 and 10.
    During training, we generate perturbations of protein sequences and learn a generator to make local edits of a base sequence to reduce its ddG value.
    At inference time, we iteratively apply the trained generator which achieves a ddG value of -5.57 after 10 iterations, more stable than the mutations seen during training.
    }
    \label{fig:fig1}
\end{figure*}

Controlled generation, \ie generating sequences $\vx$ with a desired attribute $z$, is a pervasive problem across multiple domains. In natural language processing (NLP), $z$ could represent the sentiment or the style (e.g., formality) of a sentence. In computational biology, $z$ could represent the stability, fluorescence, binding affinity, or other properties of a protein sequence. 

Occasionally, abundant supervised data of the form $(\vx,z)$ exist, such as Wikipedia domains or Gene Ontology 
categories~\cite{keskar2019ctrl,madani2020progen}, enabling direct training of a conditional generation model $p(\vx|z)$. In cases where the amount of supervised pairs available is small, it is typical to train a scorer $f(\vx)$ on this data, which maps from an input sequence to an output attribute value. %
One can then use $f(\vx)$ to annotate a large corpus for training~\cite{gehman-etal-2020-realtoxicityprompts} or directly use $f(\vx)$ during inference to guide the generation process of an unconditional
model $p(\vx)$~\cite{dathathri2019plug,yang-klein-2021-fudge}.

In this work, we focus on applications where it is necessary to generate sequences with attribute values that \textit{extrapolate} beyond the training distribution. 
For example, in biological sequence design, the problem of generating
\emph{de novo} (novel) sequences that are \emph{better} than existing natural sequences with respect to some attribute (e.g., binding affinity to a specific target) is of critical importance to drug discovery~\cite{arnold1998design,romero2009exploring,freschlin2022machine}. In creative text generation, we want to generate text that accentuates a stylistic attribute (e.g., humor) beyond simply imitating existing literature~\cite{he2019pun,lyu-etal-2021-styleptb}. 

Existing controlled generation paradigms often extrapolate poorly when the range of attribute values $z$ in the training data has limited coverage, as both $p(\vx|z)$ and the attribute scorer $f(\vx)$ may not generalize outside of the training range of the attribute. For example, consider the ACE2 stability task~\citep{chan-etal-2021-controllable} shown in Figure~\ref{fig:fig1}, where the goal is to generate mutants of the ACE2 protein that have higher stability (lower \emph{ddG} value). The training data contains sequences with \emph{ddG} values varying between $-4$ and $10$, but during inference, we want to generate more stable proteins than what we already have, \eg extrapolate to \emph{ddG} less than $-5$. Since this range of $z$ is not supported on the training data, directly fitting $p(\vx|z)$ to the training data will result in unpredictable performance for $z < -4$.

Our main assumption is that even though sequences with different target values, such as stable and unstable proteins, have distinct distributions, the process of transforming one sequence into a {\em slightly improved} version is applicable to different ranges of attribute values. For instance, in drug design, better proteins are often achieved by evolving from successive mutants, and in text generation, the sentiment can be strengthened by adding adverbs of degree. 
Therefore, we propose to the problem into a series of local improvements made to a base sequence $\vx_0$. Our intuition is that this local improvement is stable across attribute values. Thus we can learn these local edits (or mutants) on the training distribution and apply it in succession at inference time to extrapolate to new ranges of attribute values.\footnote{These iterative improvements are internal to our model and thus not analogous to rounds in directed evolution~\cite{arnold1998design}, which typically require access to a wet lab experiment (or oracle) after each round.}

As shown in Figure~\ref{fig:fig1}, to train the local editor, we synthetically generate close pairs of sequences using a masked language model~\cite{devlin-etal-2019-bert},
such that they differ marginally in attribute values. During inference, our model uses two control tags, \texttt{<inc>} for increment and \texttt{<dec>} for decrement, to locally improve a sequence in the desired direction. Increasing the number of edits on the sequence enables extrapolation. We call our approach \modelname (\modelacc).

We evaluate our approach in both the natural language and protein domains. For text generation, we generate reviews with a sentiment either more positive or negative than seen in the training data. For protein engineering, we present results on two tasks---generating mutations of the ACE2 protein that have higher stability measured by FoldX~\cite{schymkowitz2005foldx} and generating mutations of an adeno-associated virus (AAV) capsid protein~\cite{bryant2021deep} with a higher fitness value. \modelacc achieves consistent extrapolation on these three tasks, outperforming both standard methods for controlled generation such as PPLM~\cite{dathathri2019plug}
and a state-of-the-art extrapolative controlled generation method, Genhance~\cite{chan2021deep}. In particular, in the AAV task, despite seeing zero sequences that are better than the wildtype AAV sequence during training, our model is able to generate a diverse 
range of better candidates as judged by an oracle model.

\section{Related Work}
\label{related}

\subsection{Controlled Generation}
While controlled generation has been studied extensively in the literature, most of these methods do not focus on the extrapolation setting. We present an overview here situating our method and setup amongst prior work. 
\vspace{-3mm}
\paragraph{Methods using control codes}
\citet{keskar2019ctrl} and \citet{madani2020progen, madani2023large} learn a conditional sequence model $p(\vx|c)$ where $c$ is the control code,
encoding either a discrete or scalar value specifying the target attribute. 
However, these models may struggle when conditioning on unseen attribute values outside the training data range. 
Instead of conditioning on absolute target values, \citet{Lu2022QuarkCT} attempt to overcome this limitation by sampling generations from a model,  iteratively quantizing these into more fine-grained control codes and then using the highest bucket for controlled generation. 

\paragraph{Iterative editing methods}
Our approach is also related to edit-based approaches~\cite{guu2018generating,mallinson2022edit5,novak2016iterative}, and closely connected to concurrent work, \citet{welleck2022generating}, 
that samples and scores generations from a model in order to learn edits in various NLP tasks. The key distinction to our work is that we focus on extrapolation. In the setup of \citet{welleck2022generating}, the model learns by seeking feedback on all generated pairs. However, we are explicitly interested in the case where the model is required to generate sequences outside the range where it is able to obtain feedback. 


\paragraph{Latent variable models}
Another approach to achieve control is to model the attribute as a latent variable~\cite{mueller2017sequence,gligorijevic2021function,chan2021deep,chan2020cocon}.
For example, Genhance~\citep{chan2021deep} proposes to represent the latent vector as a sum of attribute-relevant and attribute-irrelevant components. They then perturb the former to achieve extrapolation with applications to both NLP and biology.
 However, latent variable models on discrete sequence data are known to suffer from stability issues. In contrast, our approach makes edits in the text space, bypassing the problem of mapping from continuous latent spaces to discrete sequences. 

\paragraph{Attribute control via a scorer model}
Another line of work~\citep{dathathri2019plug,yang-klein-2021-fudge,li2022diffusion} adds attribute information via a scorer model $p(z | \vx)$ to guide an unconditional language model $p(\vx)$ at inference time. 
Because this approach heavily relies on the scorer model which is a trained classifier, 
it is not often conducive
to extrapolation beyond the training data distribution, as we will show in our experiments.
Alternatively, one could use the classifier as a reward model for reinforcement learning~\cite{gong2019reinforcement,Angermueller2020Model-based} which suffers from similar shortcomings as the generator can exploit and amplify imperfections in the reward~\cite{amodei2016concrete,ibarz2018reward,pang2022reward}.





\subsection{Biological Sequence Design} The problem of generating \emph{de novo} sequences that improve upon natural sequences is of massive value to drug discovery, healthcare, and agriculture, as signified by the 2018  Nobel Prize in Chemistry on directed evolution~\cite{arnold1998design}. As a result, there has been a growing interest in using machine learning for this problem~\cite{yang2019machine,angermueller2020population,freschlin2022machine,ren2022proximal}. \citet{brookes2019conditioning} tackle extrapolation via a series of importance sampling distributions, in contrast to our controlled generation approach.

The iterative nature of \modelacc is internal to our modeling approach and thus not analogous to rounds in directed evolution which typically require access to an oracle (or wet lab experiment) after each round. Rather, at each round of directed evolution, \modelacc could potentially be (iteratively) run and its final output interpreted as the proposed candidates for validation.

Generating and experimentally validating novel sequences from large pretrained protein language models is also an exciting but nascent area. These approaches~\cite{madani2021deep,verkuil2022language} typically generate sequences by conditioning on broad categories or backbone structures, rather than optimizing towards a specific target attribute (e.g., stability or fluorescence) as we seek to do.

\section{Our Approach}
\label{sec:approach}

\paragraph{Problem setup}
We denote an input sequence with $\ell$ tokens as $\vx = (x_1,...,x_{\ell})$ and an attribute value as $z \in \mathbb{R}$. Here $\vx$ can represent a protein sequence of $\ell$ amino acids, where $z$ represents its stability, or a textual restaurant review of $\ell$ tokens, where $z$ corresponds to the associated sentiment score.
During training, we are typically given a large unsupervised corpus $\gD_{\unsup} = \{ \vx^{(m)} \}_{m=1}^{M_{\unsup}}$ of size $M_{\unsup}$ 
and a much smaller supervised corpus of sequences paired with attribute values, $\gD_{\suptrain} = \{ (\vx^{(m)}, z^{(m)}) \}_{m=1}^{M_{\suptrain}}$ of size $M_{\suptrain}$. 
Let $\alpha_{-}$ and $\alpha_{+}$ denote the lower and upper bound of $z$ in $\gD_{\textrm{sup-train}}$ respectively, i.e., $z \in [\alpha_{-}, \alpha_{+}]$ for all $z$ in the training examples. 
We refer to this region as the \emph{training region} of scores.

Our goal is to generate sequences that have an attribute value 
greater than (or less than) a target attribute value $z^\ast$. In particular, we aim to extrapolate beyond the training region, \ie  $z^\ast < \alpha_{-}$ or $z^\ast > \alpha_{+}$ depending on the application. We refer to these regions as the \emph{extrapolation region} of scores.

Further, we assume that we have access to 
a scorer $f_s$ that is trained on $\gD_{\textrm{sup-train}}$ to predict the attribute value of each sequence, \ie $\hat{z} = f_s(\vx)$. While $f_s$ may achieve high performance on the training region of $z$, it is not trained on data from the extrapolation region and hence it can perform poorly when scoring examples in this range. Thus $f_s$ should \emph{not} be regarded as an oracle. 

\subsection{Overview}
The core component of \modelacc is a local editor
that modifies a short span within a sequence to improve its attribute value. 

Specifically, 
it takes in an input sequence $\vx$ and
a control token $c$ that specifies whether to increase ($c=\texttt{<inc>}$) or decrease ($c=\texttt{<dec>}$) the attribute value,
and outputs an improved sequence ${\tilde\vx}$.
We model the local editor $p_{\theta}(\tilde\vx \mid \vx, c)$ using a Transformer  encoder-decoder model~\cite{vaswani2017attention}.
We train the editor by synthesizing pairs of sequences with a small difference in attribute value using masked language modeling (\Cref{sec:train_ice}). 

At inference time, 
starting with an initial sequence 
$\vx_0$,
we edit it iteratively until some stopping criteria is reached. 
Specifically, in iteration $k$, we edit the current sequence $\vx_k$ to produce $\vx_{k+1}$ by:
\vspace{-1mm}
\begin{align}
{\vx}_{k+1} \sim p_\theta(\cdot \mid \vx_k, c) 
\label{eq:iteration}
\end{align}
Each iteration is expected to move the attribute value of $\vx_k$ toward $z^\ast$. We explore different ways of selecting the best candidate at each step of the inference as well as the stopping criteria of the inference process in \Cref{sec:inference}. %

\subsection{Learning Local Edits from Perturbations}
\label{sec:train_ice}

To train the local editor, 
we perturb examples from $\gD_{\suptrain}$ %
to generate 
training pairs with a small improvement toward the target value. 

Specifically, given a sequence from the training region, $\gD_{\suptrain}$, %
we mask random tokens in it,\footnote{The specific masking strategy varies depending on the task and is specified in each of the experiment sections (\Cref{sec:sentiment}, \Cref{sec:ace2}, \Cref{sec:aav}).}
and use a masked language model to infill these 
to produce its perturbation (\Cref{fig:fig1}). 
The masked language model is trained on the unsupervised data $\gD_{\textrm{unsup}}$ such that the infill produces a valid sequence.
To ensure that we make only small improvements, 
we predict the attribute value of each sequence using the scorer $f_s$,
and retain only those pairs where the absolute difference in the attribute value is below a threshold $\delta$. 

Each pair of the original sequence and its perturbation
gives us two examples for the editor:
generating the perturbed sequence from the original sequence, and vice versa.
Recall that the editor also takes in a control token that specifies whether the edit should increase or decrease the attribute value.
For each input-output pair, we set the control code to be \texttt{<inc>} if
the attribute value of the input sequence is less than that of the output sequence measured by the scorer $f_s$,
and \texttt{<dec>} otherwise.

Given tuples of the input sequence, the output sequence, and the control code, we then train the editor $p_\theta$ on this dataset.

\subsection{Inference}
\label{sec:inference}

At inference time, we run the editor iteratively as described in Eq.~(\ref{eq:iteration}). 
\paragraph{Decoding method} During each iteration, we experiment with two different ways in which to select the best candidate out of a set of generated sequences:
\vspace{-0.3cm}
\begin{itemize}
    \itemsep0em 
    \item \textbf{Scorer-free generation}: At each iteration of \Cref{eq:iteration}, we perform generation using beam search relying on the \modelacc model likelihood to control the generation process.
    \item \textbf{Scorer-guided generation}: At each iteration, we generate a set of sequences via top-$k$ sampling, score these with $f_s$ and select the sequence assigned the highest (or lowest) score depending on the desired target value. While $f_s$ is reliable in the training region, it is unclear if the guidance provided is beneficial to the \modelacc model as it generates sequences having attribute value in the extrapolation region.
\end{itemize}

\paragraph{Stopping criteria} The objective of the task is to edit the input sequence to have an attribute value greater than (or less than) the target value $z^\ast$. However, reliably identifying when the inference process has reached $z^\ast$ is difficult as it lies in the extrapolation region. %
In this work, we run inference for a constant number of iterations. We include additional discussion on the stopping condition in \Cref{sec:stopping}.
%

\section{Experimental Setup}

We evaluate our approach on one NLP task and two protein design tasks---sentiment controlled generation (\Cref{sec:sentiment}), the ACE2 stability task (\Cref{sec:ace2}), and the AAV fitness task (\Cref{sec:aav}).

\subsection{Evaluation}
\label{sec:evaluation}
We are interested in measuring the ability of a model to successfully edit a sequence to have an attribute value greater than (or lesser than) a target value $z^*$.
In our experiments, we report the success rate or the fraction of sequences that the model is able to edit to meet this criterion as determined by an oracle model. The oracle varies based on the task and is detailed in each of the experiment sections (\Cref{sec:sentiment}, \Cref{sec:ace2}, \Cref{sec:aav}). %

\subsection{Baselines}
\label{sec:baselines}
We benchmark the performance of our method against the following baselines. 
(a) \textbf{Sampling}: A simple baseline is to directly edit sequences using a masked language model. Mirroring the synthetic data creation process from \Cref{sec:train_ice}, we mask and infill a random span within the initial sequence to change its attribute value. %
(b) \textbf{Iterative Sampling}: To ablate the contribution of the editor model in \modelacc, we replace it with a mask-and-infill editor using a masked language model; the rest of the iterative algorithm is the same as \modelacc with the \emph{Scorer-Guided} inference method. 
(c) \textbf{Genhance}: We compare to \emph{Genhance} \cite{chan2021deep}, an extrapolative baseline %
which performs controlled generation by making perturbations in a latent space learned to encode the attribute value. Increasing the size of these perturbations during inference enables extrapolation. 

For the NLP task, we compare to two additional baselines. (d) \textbf{PPLM}  \cite{dathathri2019plug} is a controlled generation method that guides the generation of an autoregressive language model at inference time using a scorer, $p(z|\vx)$. We use $f_s$ as the scorer to guide the generation. We include the baseline to evaluate if the guidance from the scorer trained on the training region allows for extrapolation. 
(e) \textbf{Score-Conditioned Generator}: We also compare to a score-conditioned model,
which generates the output sequence given the input and the target attribute value.\footnote{This baseline is similar to the methods described in \citet{jain2021empirical} and \citet{chen2021decision}.}
To train the score-conditioned model, we use the same synthetic data (\Cref{sec:train_ice}) but replace the control code with the attribute value of the output sequence measured by $f_s$ appended as a string token. At inference time, we append the desired target score and evaluate if the model generalizes to the unseen score values.\footnote{The score-conditioned baseline is trained on minimal edits and at test-time, we assess its ability to generalize to larger edits, which poses a challenge. Altering the training data to incorporate larger edits could improve the performance of this baseline however in our problem setting, we do not have pairs of sequences for the examples in $\gD_{\suptrain}$.} %

\section{Sentiment Control}
\label{sec:sentiment}

In this task, the objective is to control the sentiment associated with a short paragraph  of text (2--3 sentences). 
We use the Yelp dataset for this task \cite{zhang2015character}, which consists of 650K training examples and 50K test examples, evenly divided into sentiment scores from $1$ to $5$. We define the \emph{training region} as the range of sentiment scores from $2$ to $4$ and the \emph{extrapolation region} as the range of scores from $1$ to $2$ and $4$ to $5$. For this task, we are interested in measuring the ability of the model to extrapolate in both directions, \ie increase and decrease the associated sentiment of an example. To measure this, we report the success rate of editing the sentiment beyond the following \emph{target values}---$1.5$ and $2.5$ in the negative direction and $3.5$ and $4.5$ in the positive direction. $1.5$ and $4.5$ belong to the \emph{extrapolation region}.

\begin{table*}[ht!]
\centering
\small
\begin{tabular}{l|ccc|ccc}
\toprule
\multirow{2}{*}{\textbf{Methods}} & \multicolumn{3}{C{3cm}|}{\textbf{Targets in Training Region}}  & \multicolumn{3}{C{3cm}}{\textbf{Targets in Extrapolation Region}} \\ 
 & \multicolumn{1}{c}{\textbf{3.5}} & \multicolumn{1}{c}{\textbf{2.5}} & \textbf{Average} & \multicolumn{1}{c}{\textbf{4.5}} & \multicolumn{1}{c}{\textbf{1.5}} & \textbf{Average} \\ \midrule
Sampling  & 0.362 & 0.259 & 0.310 & 0.061 & 0.050 & 0.056 \\ 
Iterative Sampling  & 0.668 & 0.657 & 0.663 & 0.320 & 0.328 & 0.324 \\ 
Genhance   & \textbf{0.982} & 0.833 & 0.908 & 0.482 & 0.291 & 0.387 \\ 
Score-Conditioned Model & 0.780 & 0.766 & 0.773 & 0.212 & 0.217 & 0.215 \\ 
PPLM & 0.534 & 0.516 & 0.522 & 0.081 & 0.065 & 0.077 \\ \midrule
\modelacc Scorer-Free  & 0.976 & \textbf{0.918} & \textbf{0.947} & 0.446 & 0.305 & 0.376 \\ 
\modelacc w/ Scorer & 0.943 & 0.900 & 0.921 & \textbf{0.638} & \textbf{0.582} & \textbf{0.610}   \\ \bottomrule
\end{tabular}
\caption{
Results on the sentiment control task. We report the success rate measured as the fraction of examples that have a sentiment value greater than (or less than) a target score as determined by the oracle. Bold values indicate the highest rates of extrapolation. \emph{Iterative Sampling}, \emph{Genhance}, and \emph{PPLM} use the scorer for inference. \modelacc achieves the highest success rate in the extrapolation region compared to the  baselines. 
}
\label{tab:sent_results}
\end{table*}

\subsection{Implementation Details}
\paragraph{Training the scorer} We fine-tune a RoBERTa-Large model \cite{liu2019roberta} on the examples from the Yelp dataset %
in the \emph{training region} to serve as the scorer, $f_s$. The scorer is a regression model that takes in the input text and predicts its sentiment score, a real number between $2$ and $4$. \Cref{sec:expt_details} describes further training details of the scorer. 

\paragraph{Training the editor} To create the synthetic data through perturbation, we mask tokens using the strategy described in \citet{lewis2020bart} and infill these with a pre-trained BART-Large model.\footnote{The masking strategy involves sampling a location of the start of the span from a Bernoulli distribution ($p=0.8$) and then selecting the number of tokens to mask by sampling from a truncated Poisson distribution ($\lambda = 6$). The maximum span size is set to $12$. We report more variants of the masking strategy in \Cref{tab:sent_results_full} in \Cref{app:sent_results_extra}.%
} We filter the pairs created by setting the hyperparameter $\delta = 0.4$ (\Cref{sec:train_ice}).  
We fine-tune the T5-Base  model \cite{raffel2020exploring} on the synthetic training data to obtain the local editor. 
\Cref{sec:expt_details} describes further training details.

\paragraph{Inference} We run inference using both methods described in \Cref{sec:inference}. For scorer-free inference, we use beam search with a beam size of $5$.  When performing scorer-guided inference, at each iteration, we generate $5$ sequences using top-$k$ sampling with $k = 5$ and a temperature of $0.7$; we then select the best one using $f_s$. We run $10$ steps of iterative editing for both methods. 

\paragraph{Evaluation} We report results on a random subset of $1831$ examples from the test set of the Yelp dataset against all $4$ aforementioned targets.\footnote{We ensure that these examples are selected such that the sentiment value of the input text is within the \emph{training region}.} 
To evaluate whether the attribute value of the final generated sequence extrapolates beyond the \emph{training region},
we estimate  
the ground-truth sentiment scores via an oracle---a RoBERTa-Large model that is fine-tuned on the entire Yelp dataset, i.e., both the \emph{training} and \emph{extrapolation regions}.  

\paragraph{Baselines}
For sentiment control, we compare our method to \emph{Sampling}, \emph{Iterative Sampling}, \emph{Genhance}, \emph{PPLM}, and the \emph{Score-Conditioned Generator}. We use T5-Base to train the \emph{Score-Conditioned Generator} to match the \modelacc editor. The architecture of the \emph{Genhance} model is also based on T5-Base, making it comparable in size to \modelacc editor. At inference time, for each test example, we sample $50$ sequences from \emph{Genhance} and use $f_s$ to select the best one to match the total number of sequences generated by \modelacc in all iterations. For \emph{Iterative Sampling}, we generate $5$ sequences per iteration for $10$ iterations and use $f_s$ to select the best one at each iteration, the same as \modelacc.

\vspace{-2mm}
\subsection{Results}

\paragraph{\modelacc outperforms the baselines in the \emph{extrapolation region}} 
From \Cref{tab:sent_results}, we see that the \modelacc model (when guided by the scorer) strongly outperforms the baseline methods in the \emph{extrapolation region}. Even without the scorer, the \modelacc model achieves performance on par with the strongest baseline, \emph{Genhance}. \Cref{tab:example_sentiment} in \Cref{app:sent_results_extra} shows an example of increasing the sentiment associated with a sentence over multiple iterations.

\paragraph{Scorer guidance is beneficial} We observe that the scorer helps both the \emph{Iterative Sampling} baseline and \modelacc in sentiment control. \emph{Iterative Sampling} benefits from the scorer with extrapolation performance increasing to $32.4\%$ from the $5.6\%$ observed in \emph{Sampling}. %
The \modelacc success rate when guided by the scorer goes up from $37.6\%$ to $61.0\%$. We do observe that \emph{PPLM} extrapolates poorly despite using the scorer $f_s$. This highlights that $f_s$ could be more useful for guiding inference when used to rank generated sequences, as in \modelacc and \emph{Iterative Sampling}, as opposed to the conditional probabilities from $f_s$ being directly used to guide the generation, as in \emph{PPLM}.%

\begin{figure}[ht]
     \centering
     \includegraphics[width=0.45\textwidth]{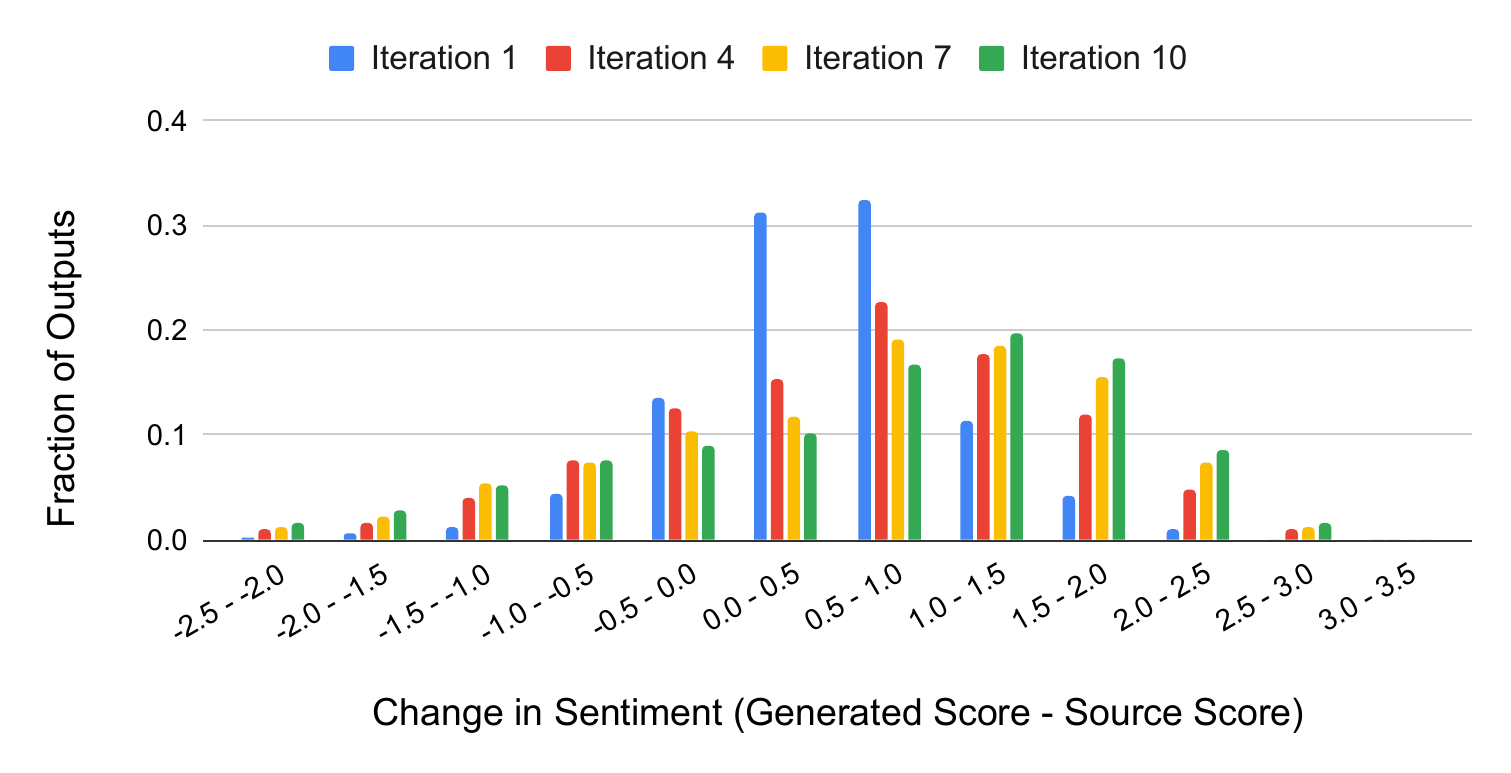}
     \caption{
     We plot the histogram of the increase in sentiment scores with respect to the initial score at every iteration of \modelacc on 100 examples. %
     As the iteration count increases, we observe that the mode of the distribution moves towards the positive side, suggesting that more examples are edited to be increasingly positive, resulting in extrapolation eventually.%
     } 
     \label{fig:iter_v_output_dist_sentiment}
\end{figure}

\vspace{-2mm}
\paragraph{What does \modelacc do in each iteration?}
To analyze how the sentiment score of the text is changed over iterations, we plot the difference between the sentiment score of the output at each iteration and that of the initial sequence.
We randomly sample 100 examples from the test set,
and use \modelacc to increase their sentiment scores.
We collect the output of \modelacc at every iteration using the scorer-free inference.
We then plot the histogram of the increase in sentiment score (with respect to the initial score) for iterations 1, 4, 7, and 10 in \Cref{fig:iter_v_output_dist_sentiment}.
%
As the iteration count increases, we observe that the increase in sentiment scores also becomes larger (i.e., the mode of the distribution is moving right),
although the editing is not always successful (the scores of a small number of outputs decrease from the initial score and fall in the negative buckets). 
Overall, this shows that \modelacc is able to increase the sentiment score on average via iterative editing.

\vspace{-2mm}
\section{Protein Design on the ACE2 dataset}
\label{sec:ace2}

Developing ways that generate more stable proteins could benefit drug discovery, as these proteins could potentially allow easier storage and have more reliable clinical effects compared to the existing proteins \cite{wang1999instability,shire2004challenges,bloom2006protein,deller2016protein,webber2016supramolecular}. The objective of this task is to generate mutants of the human angiotensin-converting enzyme 2 (ACE2) wild-type sequence\footnote{https://www.uniprot.org/uniprotkb/Q9BYF1/entry} that have higher stability. The stability value of the mutants is measured using the change in free energy from the wild-type, or \emph{ddG}, via FoldX~\cite{schymkowitz2005foldx}.\footnote{https://foldxsuite.crg.eu/} The wild-type itself has a \emph{ddG} value of zero and more negative values represent more stable mutants. This synthetic task was created in \citet{chan2021deep} and we replicate their setup. 
The proteins are represented by a sequence of $83$ amino acids out of a vocabulary of $20$ different amino acids. In order to enforce that the mutations do not diverge too widely from the wild-type, a constant span of $8$ amino acids (NTNITEEN) is kept fixed in all mutations. %
We view the \emph{training region} to be the range of \emph{ddG} values from $-4$ to $+10$. The \emph{extrapolation region} refers to \emph{ddG} values below $-4$. 
For this task, we aim to generate mutants having more negative \emph{ddG} values. We measure this by reporting the success rate of generating mutations having \emph{ddG} below target values, $z^*$, in the \emph{training region}, $-1$ and $-2.5$, and the \emph{extrapolation region}, $-5$, $-6$, and $-7$.

\begin{table*}[ht!]
\centering
\small
\begin{tabular}{l|cc|ccc}
\toprule
\multirow{2}{*}{\textbf{Methods}} & \multicolumn{2}{C{2.5cm}|}{\textbf{Targets in Training Region}} & \multicolumn{3}{C{3cm}}{\textbf{Targets in Extrapolation Region}} \\
  &  \textbf{-1}    & \textbf{-2.5}       & \textbf{-5}     & \textbf{-6}     & \textbf{-7}    \\ 
  \midrule
Sampling  & 0.033       & 0.007      & 0.000  & 0.000  & 0.000       \\
Iterative Sampling  & 0.998       & 0.954      & 0.220  & 0.079  & 0.001       \\ 
Genhance Scorer-Free & 0.570       & 0.219      & 0.021  & 0.005  & 0.001       \\
Genhance w/ Scorer    & \textbf{0.999}       & \textbf{0.978}      & 0.159  & 0.040  & 0.009       \\ \midrule
\modelacc Scorer-Free    & 0.945       & 0.598      & 0.062  & 0.017  & 0.002       \\
\modelacc w/ Scorer  & 0.998       & 0.974      & \textbf{0.361}  & \textbf{0.098}  & \textbf{0.019}       \\ \bottomrule
\end{tabular}
\caption{
Results on the ACE2 task. The objective is to generate mutants of the wild-type that have higher stability i.e. lower \emph{ddG} value. Each table cell represents the success rate of generating mutations lower than the corresponding target. Bold values indicate the highest rates of extrapolation. \modelacc achieves a higher rate of extrapolation than the reported baselines.} %
\label{tab:ace2_results}
\end{table*}

\vspace{-2mm}
\subsection{Implementation Details}
\paragraph{Training the scorer} To train $f_s$ we fine-tune {ProtBert} \cite{elnaggar2020prottrans} on the examples with \emph{ddG} values in the \emph{training region} from the dataset 
in 
\citet{chan2021deep}.
\vspace{-3mm}
\paragraph{Training the editor} We create pairs of sequences using the mask-and-infill approach from \Cref{sec:train_ice} using a pretrained Prot-T5-XL model \cite{elnaggar2020prottrans}. We sample token masks from a Bernoulli distribution with $(p=0.8)$. To filter small perturbations, we set $\delta$ to $1.5$. We then fine-tune Prot-T5-XL on this data to serve as the \modelacc editor. 

\vspace{-3mm}

\paragraph{Inference} At inference time, we start from the wild-type and generate mutations with and without the scorer, $f_s$ (\Cref{sec:inference}). When using the scorer, we sample $5$ sequences at each step, select the best one using $f_s$, and repeat the process for $10$ iterations. For scorer-free inference, we generate sequences with beam size of $5$ for $10$ iterations.\footnote{We present further analysis on the variation in performance based on the hyperparameters of generation in \Cref{app:hyperparameter_experiment}.} %

\vspace{-3mm}

\paragraph{Evaluation} In the ACE2 task, we are interested in generating mutants that have a lower \emph{ddG} value. So we generate $10,000$ mutants of the wild-type from each model and report the success rate of generating mutants that have a \emph{ddG} value lower than each of the task targets using FoldX as the oracle. We match the FoldX evaluation parameters from \citet{chan2021deep} to evaluate the mutations. %
We also report the average score of the Top-100 and Top-1000 mutants as determined by the oracle to evaluate the quality of the top candidates in the library of 10,000 produced by each model. 

\vspace{-3mm}

\paragraph{Baselines} We compare our approach against \emph{Sampling}, \emph{Iterative Sampling}, and \emph{Genhance}.\footnote{The ACE2 task requires generating mutants of a specific wild-type. Pretrained autoregressive language models in the protein domain cannot generate mutants directly, only continuing sequences. As a result, sequence-to-sequence models are more appropriate for this task. Hence, {\em PPLM}, which relies on an autoregressive model, is not included as a baseline. %
Also, we do not include the \emph{Score-Conditioned Generator} baseline as the vocabulary of Prot-T5-XL tokenizer solely consists of amino acids, thus it cannot accept the output score as a token along with the input. %
} %
For \emph{Genhance}, we report results from the model released by \citet{chan2021deep} on $10,000$ mutants generated with and without the scorer. This model is based on Prot-T5-XL as well making it directly comparable to the \modelacc model. 
For \emph{Iterative Sampling}, we generate $5$ sequences per iteration for $10$ iterations.

\subsection{Results}
\begin{figure*}[ht!]
     \centering
     \begin{subfigure}[b]{0.475\textwidth}
   \centering
   \includegraphics[width=\textwidth]{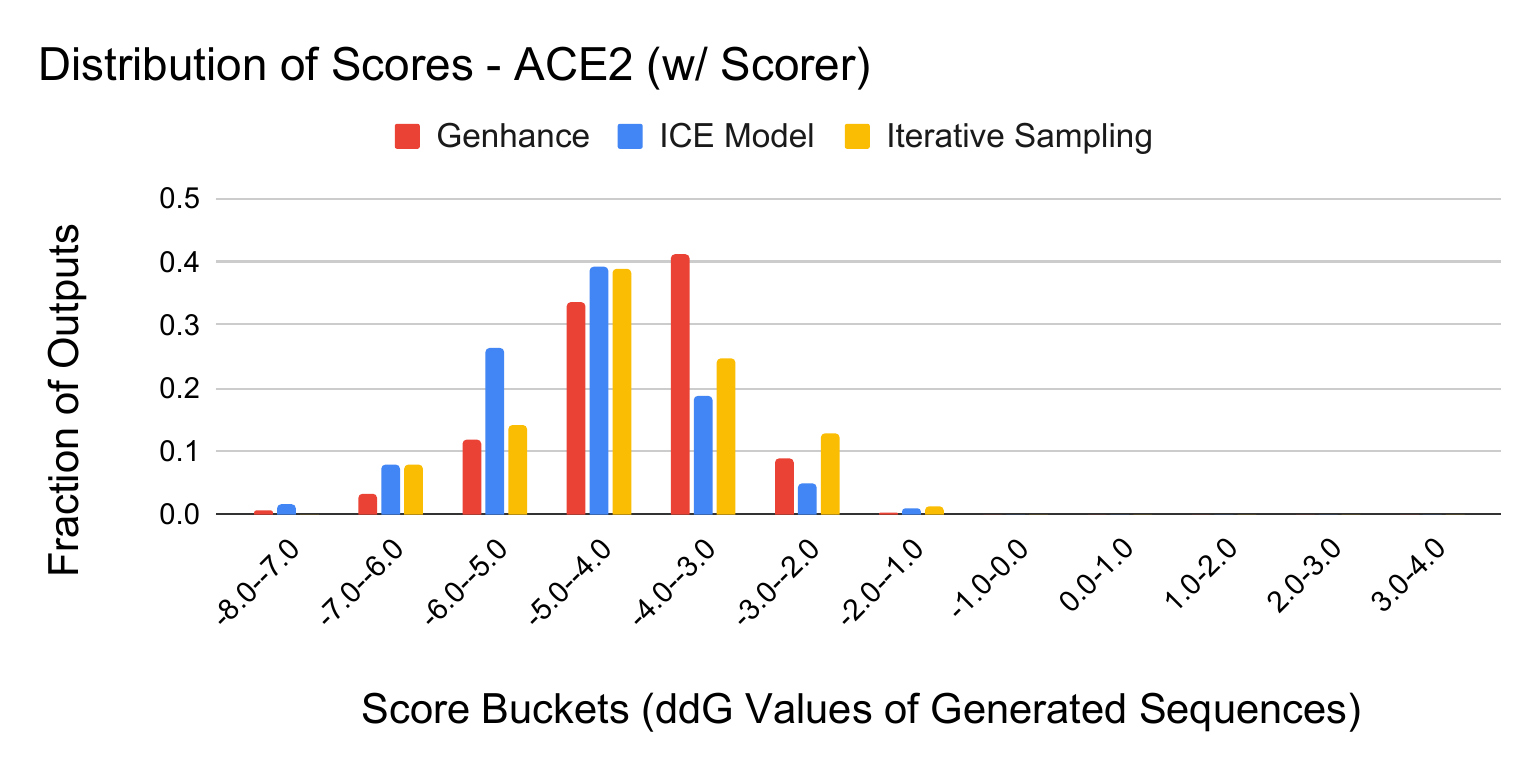}
   \caption{}
   \label{fig:ace2-dist-w-disc}
     \end{subfigure}
     \hfill
     \begin{subfigure}[b]{0.475\textwidth}
   \centering
   \includegraphics[width=\textwidth]{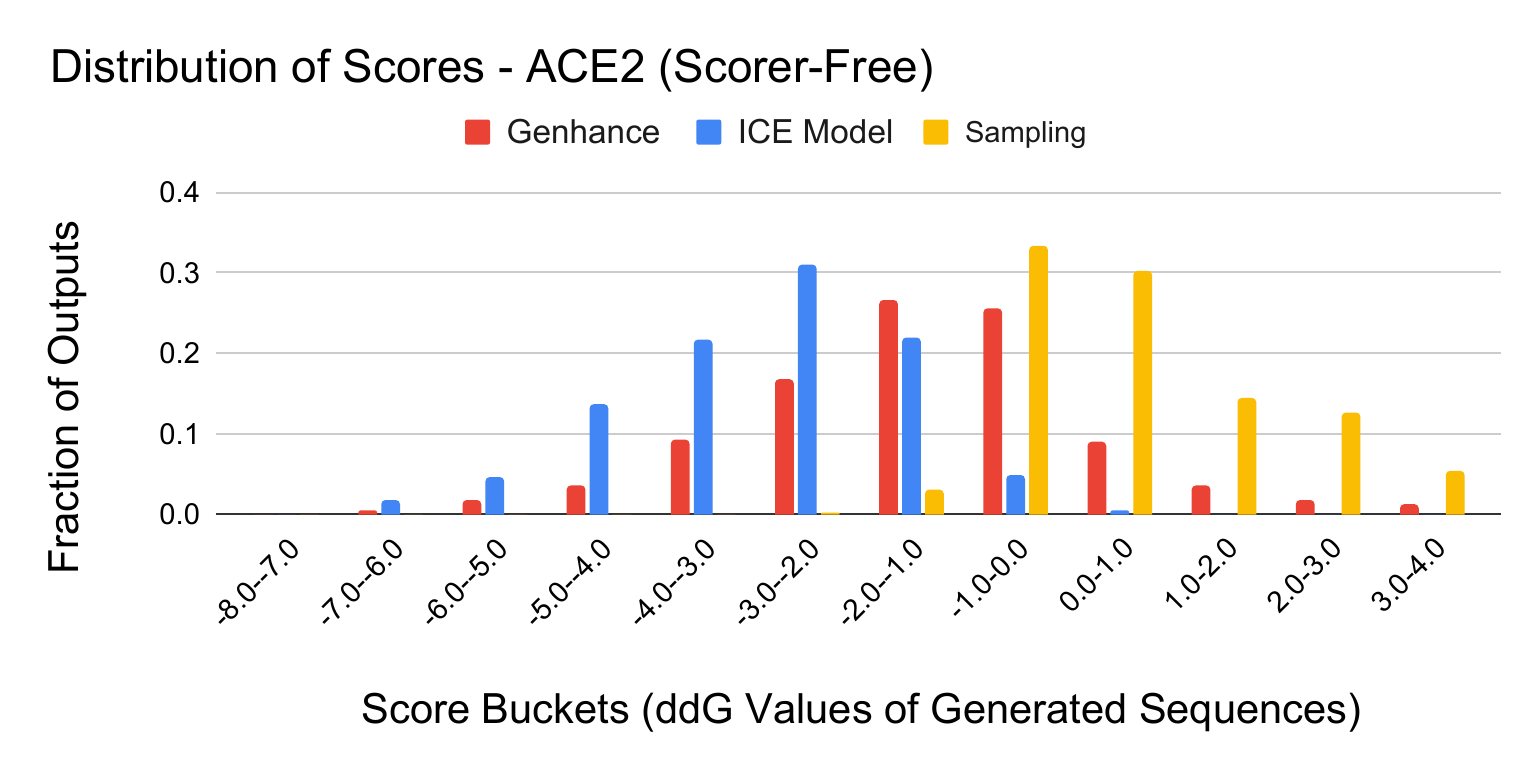}
   \caption{}
   \label{fig:ace2-dist-wo-disc}
     \end{subfigure}
     \caption{
     Histograms of \emph{ddG} scores (lower is better) of the final mutations generated by \modelacc and the baselines on the ACE2 task. 
     \modelacc generates higher quality mutations than the baselines both with (\Cref{fig:ace2-dist-w-disc}) and without the scorer (\Cref{fig:ace2-dist-wo-disc}) guiding the inference. Further, the scorer significantly improves performance for all methods.}   %
     \label{fig:output_dist_ace2}
\end{figure*}

\paragraph{\modelacc outperforms baselines on extrapolation} \Cref{tab:ace2_results} shows that \modelacc consistently outperforms \emph{Genhance}, \emph{Sampling}, and \emph{Iterative Sampling} on all extrapolation targets. In addition, from \Cref{tab:ace2_topk_scores}, we see that \modelacc achieves a lower average \emph{ddG} on the Top-100 and Top-1000 sequences. Interestingly, while \emph{Iterative Sampling} achieves higher extrapolation rates than \emph{Genhance} (\Cref{tab:ace2_results}), \emph{Genhance} achieves a better average score on the Top-1000 and Top-100 subsets (\Cref{tab:ace2_topk_scores}) indicating that \emph{Genhance} produces a smaller number of slightly more stable mutants (though still outperformed by \modelacc).
\vspace{-2mm}
\paragraph{The scorer is  valuable for all models in ACE2} In this task, we begin the generation from the wild-type (\emph{ddG} score of zero) and the scorer, $f_s$, reliably guides the generation process until the score of $-5$. As a result, we see that all the methods strongly benefit from using the scorer (\Cref{tab:ace2_results}). In \Cref{fig:output_dist_ace2}, we plot the histogram of scores of the generated mutations from \modelacc and the reported baselines. %
From \Cref{fig:ace2-dist-w-disc}, we see that the peaks of the distribution of scores for all models move in the negative direction to be centered closer to $-5$ as compared to \Cref{fig:ace2-dist-wo-disc} highlighting the value of the scorer. We do however note that our approach is able to achieve some extrapolation even in the scorer-free regime, far outperforming \emph{Sampling} and achieving extrapolation at a higher rate than \emph{Genhance}.\\

\begin{table}[ht!]
\centering
\small
\begin{tabular}{r|c|c|c}
\toprule
\textbf{Library Size} & \makecell[c]{\textbf{Iterative} \\ \textbf{Sampling}} & \textbf{Genhance} & \textbf{\modelacc} \\ \midrule
\multicolumn{1}{r|}{\makecell[c]{All 10k}}     & -4.326   & -4.086   & \textbf{-4.660}    \\
\multicolumn{1}{r|}{\makecell[c]{Top 1k}}  & -5.866   & -6.030   & \textbf{-6.575}    \\
\multicolumn{1}{r|}{\makecell[c]{Top 100}} & -6.413   & -7.354   & \textbf{-7.938}    \\ \bottomrule
\end{tabular}
\caption{
Average \emph{ddG} values (lower is better) of mutations generated from \emph{Iterative Sampling}, \emph{Genhance}, and \modelacc (each with the scorer). We report the average score of all $10000$ mutations, the top $1000$, and the top $100$ as determined by the oracle. Bold values are the lowest average \emph{ddG} value. \modelacc generates the most stable mutations. 
}
\label{tab:ace2_topk_scores}
\end{table}

\vspace{-6mm}
\section{Protein Design on the AAV dataset}
\label{sec:aav}
\begin{table*}[ht!]
\centering
\small
\begin{tabular}{L{4cm}|c|ccc}
\toprule
\multicolumn{1}{l|}{\multirow{2}{*}{\textbf{Methods}}} & \multicolumn{1}{C{2.5cm}|}{\textbf{Targets in Training Region}} & \multicolumn{3}{C{3cm}}{\textbf{Targets in Extrapolation Region}} \\
\multicolumn{1}{l|}{}         & \textbf{-1}        & \textbf{0}   & \textbf{1}   & \textbf{2}  \\ \midrule
Sampling   & 0.058      & 0.018  & 0.011  & 0.000       \\
Iterative Sampling  w/ Scorer     & \textbf{0.524}      & 0.064  & 0.017  & 0.000       \\ \midrule
\modelacc Scorer-Free      & 0.481      & 0.188  & 0.033  & 0.001       \\
\modelacc w/ Scorer      & 0.521      & \textbf{0.223}  & \textbf{0.036}  & \textbf{0.002}       \\ \bottomrule
\end{tabular}
\caption{Results on the AAV task. The objective is to generate mutations of the source protein that have a higher fitness value. We report the success rate of generating mutations with fitness values higher than the corresponding targets. Bold values indicate the highest extrapolation rates. \modelacc achieves a higher rate of extrapolation than the baselines. } 
\label{tab:aav_results}
\end{table*}


The AAV dataset~\citep{bryant2021deep}  aims to study the fitness landscape of an adeno-associated virus (AAV) capsid protein that is a key component of gene therapy~\cite{russell2017efficacy}. Our goal is to obtain mutants of the AAV-2 wild type sequence\footnote{https://www.uniprot.org/uniprotkb/P03135/entry} that have a higher fitness value. We use the splits proposed by the FLIP benchmark~\cite{dallago2022flip} for our experiments. Each mutant is a sequence of length varying from $734$ to $750$. Mutations are made on the wild-type sequence between indices $561$ and $588$. We use the provided \emph{low-vs-high} split of the dataset to demarcate the \emph{training region} and \emph{extrapolation region}. The \emph{training region} corresponds to fitness values below zero and the \emph{extrapolation region} corresponds to positive fitness values.
At inference time, the generation process begins at the wild-type, with a fitness score of zero, and the model is expected to 
generate mutants that have a positive fitness score. We evaluate performance against target values, $z^*$ in the \emph{training region}, $-1$, and in the \emph{extrapolation region}, $0$, $1$, and $2$. 
\vspace{-1mm}
\subsection{Implementation Details}
\paragraph{Training the scorer}  The scorer, $f_s$, is a CNN model trained on the examples in the \emph{training region}. The architecture and hyperparameters for the CNN were chosen based on 
the FLIP benchmark.\footnote{On the low-vs-high split, the train correlation of the scorer is $0.82$ and the test correlation is $0.34$. This matches the best test correlation on this split obtained as part of the benchmark.} The scorer accepts a string corresponding to the proteins and outputs a floating-point fitness value.
\vspace{-2mm}
\paragraph{Training the editor} We create pairs to train the \modelacc model by 
following the same strategy as in ACE2. We use the Prot-T5-XL \cite{elnaggar2020prottrans} model to infill masks in the mutable region and score pairs with the scorer, $f_s$, to create the editor training data.\footnote{We again set the hyperparameter $\delta$ to $1.5$.} We then fine-tune Prot-T5-XL on this dataset. Since the length of the mutants is greater than the sequence length limit of Prot-T5-XL, we truncate them from the start to the last $512$ tokens, which always contain the entire mutable region of the protein.
\vspace{-3mm}
\paragraph{Inference} We start from the wild-type and run inference on the \modelacc model as per \Cref{sec:inference}. When using the scorer, we sample $5$ generations, score them with $f_s$, select the best one, and repeat for $10$ iterations. For the scorer-free setup, we generate with a beam size of $5$ for $10$ iterations. 
\vspace{-2mm}
\paragraph{Evaluation} 
We generate $10,000$ mutants with each method and report the success rate of generating mutations that are above the target scores, $z^*$. In lieu of a wet-lab experiment, we obtain fitness scores for each generated sequence via an oracle model, which is a CNN trained on the \emph{sampled} (i.i.d.) split
of the AAV dataset.\footnote{We select the CNN architecture as it has the highest spearman correlation with the gold fitness values on the benchmark \cite{dallago2022flip}. The model obtains a train spearman correlation of $0.93$ and a test correlation of $0.92$ on this split.} This was chosen as the examples from the \emph{sampled} split span fitness values across both the \emph{training region} and \emph{extrapolation region}.


\vspace{-3mm}
\paragraph{Baselines} We compare our approach to the \emph{Sampling} and \emph{Iterative Sampling} baselines.\footnote{As mentioned earlier, the \emph{PPLM} and \emph{Score-Conditioned Generator} baselines are not well suited for the protein tasks.  
} %
%

\vspace{-2mm}
\begin{table}[ht]
\centering
\small
\begin{tabular}{r|c|c|c|c}
\toprule
\makecell{\textbf{Library}\\\textbf{Size}} & \makecell{\textbf{Samp-}\\\textbf{ling}} & \makecell[c]{\textbf{Iterative} \\ \textbf{Sampling}} & \makecell[c]{\textbf{\modelacc} \\ \textbf{Scorer-} \\ \textbf{Free}} & \makecell[c]{\textbf{\modelacc} \\ \textbf{w/ Scorer}} \\ \midrule
\multicolumn{1}{r|}{\makecell[c]{All 10k}}  & -3.450   & -1.390   & -1.150 & \textbf{-1.040}   \\
\multicolumn{1}{r|}{\makecell[c]{Top 1k}}  & -0.567  & -0.584  & 0.403 &  \textbf{0.918}    \\
\multicolumn{1}{r|}{\makecell[c]{Top 100}} & 1.605 & 1.550 & 1.452 & \textbf{1.750}   \\ \hline
\end{tabular}
\caption{Average fitness values (higher is better) of mutations generated from \emph{Sampling}, \emph{Iterative Sampling}, and \modelacc. We report the average score of all $10000$ mutations, the average of the top $1000$, and the top $100$ as determined by the oracle. Bold values are the highest average fitness value. \modelacc generates the highest quality mutations.} 
\label{tab:aav_results_topk}
\end{table}

\begin{figure}[ht]
     \centering
    \includegraphics[width=0.45\textwidth]{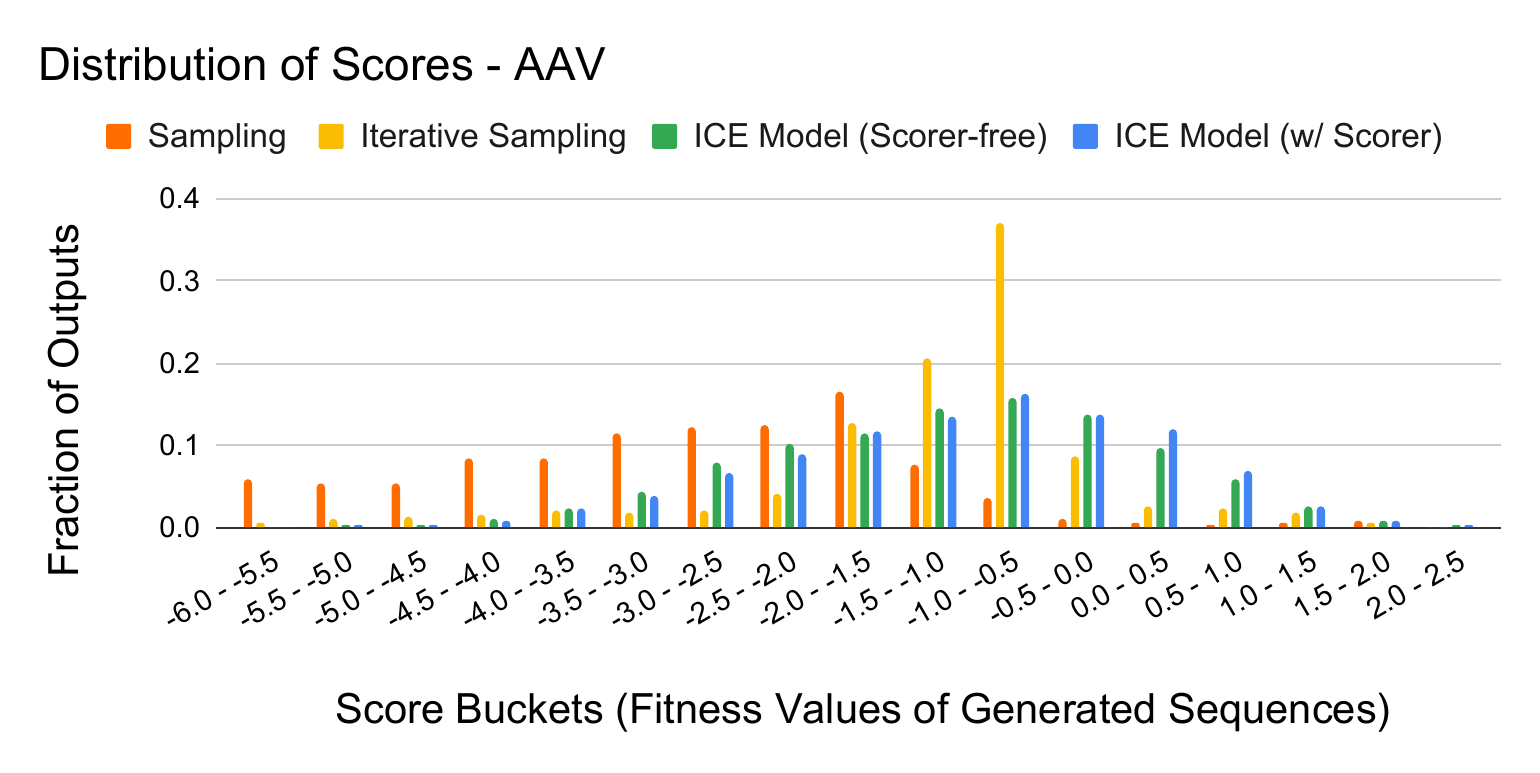}
     \caption{Histogram of fitness values of mutants generated by each approach on the AAV Task (higher scores are better). \modelacc generates outperforms \emph{Sampling} and \emph{Iterative Sampling}.
     \label{fig:aav-dist}}
     \label{fig:output_dist_aav}
\end{figure}

\subsection{Results}

\paragraph{\modelacc model extrapolates better than \emph{Iterative Sampling}} From \Cref{tab:aav_results}, we see that \modelacc with \emph{Scorer-Free} and \emph{Scorer-Guided} inference achieves a higher success rate of extrapolation than \emph{Sampling} and \emph{Iterative Sampling} respectively. 
We also observe that \modelacc with \emph{Scorer-Guided} inference achieves a higher average fitness score than the baselines on the total library of $10000$ mutations as well as the subsets of Top-$100$ and Top-$1000$ mutations generated by each method.
Lastly, it is also desirable to generate a library of mutations that not only achieves high fitness values but also exhibits diversity \cite{calcedo2009worldwide}. We observe that \modelacc generates diverse and high-quality mutations by examining the edit distance between the mutations generated and the wild-type in \Cref{app:aav_results_extra}. 

\vspace{-2mm}
\paragraph{The scorer is less effective on AAV} From \Cref{tab:aav_results}, we see that the performance of both methods on the \emph{training region} and \emph{extrapolation region} targets when using the scorer improves only marginally over the scorer-free setups. The distribution of scores (\Cref{fig:aav-dist}) also shows a similar trend. 
We see that, for both methods, the mode of the distribution of scores is within the \emph{training region} itself, close to the boundary of the \emph{extrapolation region} (\Cref{fig:aav-dist}). %
The distribution for  \modelacc  is much flatter, which is why it achieves higher extrapolation success rates compared to \emph{Iterative Sampling}. Since the generation process begins at the edge of the \emph{training region} (zero), we expect the scorer to not offer much reliable guidance in AAV. 

%

\section{Conclusion}
\label{sec:conclusion}

We presented \modelname (\modelacc), an iterative approach to extrapolative controlled generation.
Our method considerably outperforms existing approaches to controllable generation and more complex extrapolative techniques on both NLP and protein design tasks. Potential future directions include extending the iterative approach to multiple attributes to generate sequences that compose them in novel ways, training scorers that generalize to the extrapolation region, and improving our synthetic data creation techniques by incorporating additional domain knowledge.

\section*{Acknowledgements}
We thank David Belanger, Lucy Colwell, and Nitish Joshi for their valuable discussion and feedback during the course of the project. This work was undertaken as part of the Google Research Collabs program. This work is also supported by the Samsung Advanced Institute of Technology (Next Generation Deep Learning: From Pattern Recognition to AI), the National Science Foundation under Grant No. 1922658, and a gift from AWS AI.

\clearpage
\bibliography{anthology,custom}

\begin{thebibliography}{53}
\providecommand{\natexlab}[1]{#1}
\providecommand{\url}[1]{\texttt{#1}}
\expandafter\ifx\csname urlstyle\endcsname\relax
  \providecommand{\doi}[1]{doi: #1}\else
  \providecommand{\doi}{doi: \begingroup \urlstyle{rm}\Url}\fi

\bibitem[Amodei et~al.(2016)Amodei, Olah, Steinhardt, Christiano, Schulman, and
  Man{\'e}]{amodei2016concrete}
Amodei, D., Olah, C., Steinhardt, J., Christiano, P., Schulman, J., and
  Man{\'e}, D.
\newblock Concrete problems in ai safety.
\newblock \emph{arXiv preprint arXiv:1606.06565}, 2016.

\bibitem[Angermueller et~al.(2020{\natexlab{a}})Angermueller, Belanger, Gane,
  Mariet, Dohan, Murphy, Colwell, and Sculley]{angermueller2020population}
Angermueller, C., Belanger, D., Gane, A., Mariet, Z., Dohan, D., Murphy, K.,
  Colwell, L., and Sculley, D.
\newblock Population-based black-box optimization for biological sequence
  design.
\newblock In \emph{International Conference on Machine Learning}, pp.\
  324--334. PMLR, 2020{\natexlab{a}}.

\bibitem[Angermueller et~al.(2020{\natexlab{b}})Angermueller, Dohan, Belanger,
  Deshpande, Murphy, and Colwell]{Angermueller2020Model-based}
Angermueller, C., Dohan, D., Belanger, D., Deshpande, R., Murphy, K., and
  Colwell, L.
\newblock Model-based reinforcement learning for biological sequence design.
\newblock In \emph{International Conference on Learning Representations},
  2020{\natexlab{b}}.
\newblock URL \url{https://openreview.net/forum?id=HklxbgBKvr}.

\bibitem[Arnold(1998)]{arnold1998design}
Arnold, F.~H.
\newblock Design by directed evolution.
\newblock \emph{Accounts of Chemical Research}, 31\penalty0 (3):\penalty0
  125--131, 1998.

\bibitem[Bloom et~al.(2006)Bloom, Labthavikul, Otey, and
  Arnold]{bloom2006protein}
Bloom, J.~D., Labthavikul, S.~T., Otey, C.~R., and Arnold, F.~H.
\newblock Protein stability promotes evolvability.
\newblock \emph{Proceedings of the National Academy of Sciences}, 103\penalty0
  (15):\penalty0 5869--5874, 2006.
\newblock \doi{10.1073/pnas.0510098103}.
\newblock URL \url{https://www.pnas.org/doi/abs/10.1073/pnas.0510098103}.

\bibitem[Brookes et~al.(2019)Brookes, Park, and
  Listgarten]{brookes2019conditioning}
Brookes, D., Park, H., and Listgarten, J.
\newblock Conditioning by adaptive sampling for robust design.
\newblock In \emph{International conference on machine learning}, pp.\
  773--782. PMLR, 2019.

\bibitem[Bryant et~al.(2021)Bryant, Bashir, Sinai, Jain, Ogden, Riley, Church,
  Colwell, and Kelsic]{bryant2021deep}
Bryant, D.~H., Bashir, A., Sinai, S., Jain, N.~K., Ogden, P.~J., Riley, P.~F.,
  Church, G.~M., Colwell, L.~J., and Kelsic, E.~D.
\newblock Deep diversification of an aav capsid protein by machine learning.
\newblock \emph{Nature Biotechnology}, 39\penalty0 (6):\penalty0 691--696,
  2021.

\bibitem[Calcedo et~al.(2009)Calcedo, Vandenberghe, Gao, Lin, and
  Wilson]{calcedo2009worldwide}
Calcedo, R., Vandenberghe, L.~H., Gao, G., Lin, J., and Wilson, J.~M.
\newblock Worldwide epidemiology of neutralizing antibodies to adeno-associated
  viruses.
\newblock \emph{The Journal of infectious diseases}, 199\penalty0 (3):\penalty0
  381--390, 2009.

\bibitem[Chan et~al.(2021{\natexlab{a}})Chan, Madani, Krause, and
  Naik]{chan2021deep}
Chan, A., Madani, A., Krause, B., and Naik, N.
\newblock Deep extrapolation for attribute-enhanced generation.
\newblock In Beygelzimer, A., Dauphin, Y., Liang, P., and Vaughan, J.~W.
  (eds.), \emph{Advances in Neural Information Processing Systems},
  2021{\natexlab{a}}.
\newblock URL \url{https://openreview.net/forum?id=NCDMYD2y5kK}.

\bibitem[Chan et~al.(2021{\natexlab{b}})Chan, Ong, Pung, Zhang, and
  Fu]{chan2020cocon}
Chan, A., Ong, Y.-S., Pung, B., Zhang, A., and Fu, J.
\newblock Cocon: A self-supervised approach for controlled text generation.
\newblock In \emph{International Conference on Learning Representations},
  2021{\natexlab{b}}.
\newblock URL \url{https://openreview.net/forum?id=VD_ozqvBy4W}.

\bibitem[Chan et~al.(2021{\natexlab{c}})Chan, Wang, and
  King]{chan-etal-2021-controllable}
Chan, H.~P., Wang, L., and King, I.
\newblock Controllable summarization with constrained {M}arkov decision
  process.
\newblock \emph{Transactions of the Association for Computational Linguistics},
  9:\penalty0 1213--1232, 2021{\natexlab{c}}.
\newblock \doi{10.1162/tacl_a_00423}.
\newblock URL \url{https://aclanthology.org/2021.tacl-1.72}.

\bibitem[Chen et~al.(2021)Chen, Lu, Rajeswaran, Lee, Grover, Laskin, Abbeel,
  Srinivas, and Mordatch]{chen2021decision}
Chen, L., Lu, K., Rajeswaran, A., Lee, K., Grover, A., Laskin, M., Abbeel, P.,
  Srinivas, A., and Mordatch, I.
\newblock Decision transformer: Reinforcement learning via sequence modeling.
\newblock In Beygelzimer, A., Dauphin, Y., Liang, P., and Vaughan, J.~W.
  (eds.), \emph{Advances in Neural Information Processing Systems}, 2021.
\newblock URL \url{https://openreview.net/forum?id=a7APmM4B9d}.

\bibitem[Dallago et~al.(2021)Dallago, Mou, Johnston, Wittmann, Bhattacharya,
  Goldman, Madani, and Yang]{dallago2022flip}
Dallago, C., Mou, J., Johnston, K.~E., Wittmann, B., Bhattacharya, N., Goldman,
  S., Madani, A., and Yang, K.~K.
\newblock {FLIP}: Benchmark tasks in fitness landscape inference for proteins.
\newblock In \emph{Thirty-fifth Conference on Neural Information Processing
  Systems Datasets and Benchmarks Track}, 2021.
\newblock URL \url{https://openreview.net/forum?id=p2dMLEwL8tF}.

\bibitem[Dathathri et~al.(2020)Dathathri, Madotto, Lan, Hung, Frank, Molino,
  Yosinski, and Liu]{dathathri2019plug}
Dathathri, S., Madotto, A., Lan, J., Hung, J., Frank, E., Molino, P., Yosinski,
  J., and Liu, R.
\newblock Plug and play language models: A simple approach to controlled text
  generation.
\newblock In \emph{International Conference on Learning Representations}, 2020.
\newblock URL \url{https://openreview.net/forum?id=H1edEyBKDS}.

\bibitem[Deller et~al.(2016)Deller, Kong, and Rupp]{deller2016protein}
Deller, M.~C., Kong, L., and Rupp, B.
\newblock Protein stability: a crystallographer's perspective.
\newblock \emph{Acta Crystallographica Section F: Structural Biology
  Communications}, 72\penalty0 (2):\penalty0 72--95, 2016.

\bibitem[Devlin et~al.(2019)Devlin, Chang, Lee, and
  Toutanova]{devlin-etal-2019-bert}
Devlin, J., Chang, M.-W., Lee, K., and Toutanova, K.
\newblock {BERT}: Pre-training of deep bidirectional transformers for language
  understanding.
\newblock In \emph{Proceedings of the 2019 Conference of the North {A}merican
  Chapter of the Association for Computational Linguistics: Human Language
  Technologies, Volume 1 (Long and Short Papers)}, pp.\  4171--4186,
  Minneapolis, Minnesota, June 2019. Association for Computational Linguistics.
\newblock \doi{10.18653/v1/N19-1423}.
\newblock URL \url{https://aclanthology.org/N19-1423}.

\bibitem[Elnaggar et~al.(2021)Elnaggar, Heinzinger, Dallago, Rehawi, Wang,
  Jones, Gibbs, Feher, Angerer, Steinegger, et~al.]{elnaggar2020prottrans}
Elnaggar, A., Heinzinger, M., Dallago, C., Rehawi, G., Wang, Y., Jones, L.,
  Gibbs, T., Feher, T., Angerer, C., Steinegger, M., et~al.
\newblock Prottrans: Toward understanding the language of life through
  self-supervised learning.
\newblock \emph{IEEE transactions on pattern analysis and machine
  intelligence}, 44\penalty0 (10):\penalty0 7112--7127, 2021.

\bibitem[Freschlin et~al.(2022)Freschlin, Fahlberg, and
  Romero]{freschlin2022machine}
Freschlin, C.~R., Fahlberg, S.~A., and Romero, P.~A.
\newblock Machine learning to navigate fitness landscapes for protein
  engineering.
\newblock \emph{Current Opinion in Biotechnology}, 75:\penalty0 102713, 2022.

\bibitem[Gehman et~al.(2020)Gehman, Gururangan, Sap, Choi, and
  Smith]{gehman-etal-2020-realtoxicityprompts}
Gehman, S., Gururangan, S., Sap, M., Choi, Y., and Smith, N.~A.
\newblock {R}eal{T}oxicity{P}rompts: Evaluating neural toxic degeneration in
  language models.
\newblock In \emph{Findings of the Association for Computational Linguistics:
  EMNLP 2020}, pp.\  3356--3369, Online, November 2020. Association for
  Computational Linguistics.
\newblock \doi{10.18653/v1/2020.findings-emnlp.301}.
\newblock URL \url{https://aclanthology.org/2020.findings-emnlp.301}.

\bibitem[Gligorijevi{\'c} et~al.(2021)Gligorijevi{\'c}, Berenberg, Ra, Watkins,
  Kelow, Cho, and Bonneau]{gligorijevic2021function}
Gligorijevi{\'c}, V., Berenberg, D., Ra, S., Watkins, A., Kelow, S., Cho, K.,
  and Bonneau, R.
\newblock Function-guided protein design by deep manifold sampling.
\newblock \emph{bioRxiv}, 2021.
\newblock \doi{10.1101/2021.12.22.473759}.
\newblock URL
  \url{https://www.biorxiv.org/content/early/2021/12/23/2021.12.22.473759}.

\bibitem[Gong et~al.(2019)Gong, Bhat, Wu, Xiong, and
  Hwu]{gong2019reinforcement}
Gong, H., Bhat, S., Wu, L., Xiong, J., and Hwu, W.-M.
\newblock Reinforcement learning based text style transfer without parallel
  training corpus.
\newblock In \emph{Proceedings of the 2019 Conference of the North American
  Chapter of the Association for Computational Linguistics: Human Language
  Technologies, Volume 1 (Long and Short Papers)}, pp.\  3168--3180, 2019.

\bibitem[Guu et~al.(2018)Guu, Hashimoto, Oren, and Liang]{guu2018generating}
Guu, K., Hashimoto, T.~B., Oren, Y., and Liang, P.
\newblock Generating sentences by editing prototypes.
\newblock \emph{Transactions of the Association for Computational Linguistics},
  6:\penalty0 437--450, 2018.

\bibitem[He et~al.(2019)He, Peng, and Liang]{he2019pun}
He, H., Peng, N., and Liang, P.
\newblock Pun generation with surprise.
\newblock In \emph{North American Chapter of the Association for Computational
  Linguistics (NAACL)}, 2019.

\bibitem[Ibarz et~al.(2018)Ibarz, Leike, Pohlen, Irving, Legg, and
  Amodei]{ibarz2018reward}
Ibarz, B., Leike, J., Pohlen, T., Irving, G., Legg, S., and Amodei, D.
\newblock Reward learning from human preferences and demonstrations in atari.
\newblock \emph{Advances in neural information processing systems}, 31, 2018.

\bibitem[Jain \& Berg-Kirkpatrick(2021)Jain and
  Berg-Kirkpatrick]{jain2021empirical}
Jain, A. and Berg-Kirkpatrick, T.
\newblock An empirical study of extrapolation in text generation with scalar
  control.
\newblock \emph{arXiv preprint arXiv:2104.07910}, 2021.

\bibitem[Keskar et~al.(2019)Keskar, McCann, Varshney, Xiong, and
  Socher]{keskar2019ctrl}
Keskar, N.~S., McCann, B., Varshney, L.~R., Xiong, C., and Socher, R.
\newblock {CTRL}: A conditional transformer language model for controllable
  generation.
\newblock \emph{arXiv preprint arXiv:1909.05858}, 2019.

\bibitem[Lewis et~al.(2020)Lewis, Liu, Goyal, Ghazvininejad, Mohamed, Levy,
  Stoyanov, and Zettlemoyer]{lewis2020bart}
Lewis, M., Liu, Y., Goyal, N., Ghazvininejad, M., Mohamed, A., Levy, O.,
  Stoyanov, V., and Zettlemoyer, L.
\newblock Bart: Denoising sequence-to-sequence pre-training for natural
  language generation, translation, and comprehension.
\newblock In \emph{Proceedings of the 58th Annual Meeting of the Association
  for Computational Linguistics}, pp.\  7871--7880, 2020.

\bibitem[Li et~al.(2022)Li, Thickstun, Gulrajani, Liang, and
  Hashimoto]{li2022diffusion}
Li, X.~L., Thickstun, J., Gulrajani, I., Liang, P., and Hashimoto, T.
\newblock Diffusion-{LM} improves controllable text generation.
\newblock In \emph{Advances in Neural Information Processing Systems}, 2022.
\newblock URL \url{https://openreview.net/forum?id=3s9IrEsjLyk}.

\bibitem[Liu et~al.(2019)Liu, Ott, Goyal, Du, Joshi, Chen, Levy, Lewis,
  Zettlemoyer, and Stoyanov]{liu2019roberta}
Liu, Y., Ott, M., Goyal, N., Du, J., Joshi, M., Chen, D., Levy, O., Lewis, M.,
  Zettlemoyer, L., and Stoyanov, V.
\newblock Roberta: A robustly optimized bert pretraining approach.
\newblock \emph{arXiv preprint arXiv:1907.11692}, 2019.

\bibitem[Lu et~al.(2022)Lu, Welleck, Hessel, Jiang, Qin, West, Ammanabrolu, and
  Choi]{Lu2022QuarkCT}
Lu, X., Welleck, S., Hessel, J., Jiang, L., Qin, L., West, P., Ammanabrolu, P.,
  and Choi, Y.
\newblock Quark: Controllable text generation with reinforced unlearning.
\newblock \emph{Advances in neural information processing systems},
  35:\penalty0 27591--27609, 2022.

\bibitem[Lyu et~al.(2021)Lyu, Liang, Pham, Hovy, P{\'o}czos, Salakhutdinov, and
  Morency]{lyu-etal-2021-styleptb}
Lyu, Y., Liang, P.~P., Pham, H., Hovy, E., P{\'o}czos, B., Salakhutdinov, R.,
  and Morency, L.-P.
\newblock {S}tyle{PTB}: A compositional benchmark for fine-grained controllable
  text style transfer.
\newblock In \emph{Proceedings of the 2021 Conference of the North American
  Chapter of the Association for Computational Linguistics: Human Language
  Technologies}, pp.\  2116--2138, Online, June 2021. Association for
  Computational Linguistics.
\newblock \doi{10.18653/v1/2021.naacl-main.171}.
\newblock URL \url{https://aclanthology.org/2021.naacl-main.171}.

\bibitem[Madani et~al.(2020)Madani, McCann, Naik, Keskar, Anand, Eguchi, Huang,
  and Socher]{madani2020progen}
Madani, A., McCann, B., Naik, N., Keskar, N.~S., Anand, N., Eguchi, R.~R.,
  Huang, P.-S., and Socher, R.
\newblock {ProGen}: Language modeling for protein generation.
\newblock \emph{arXiv preprint arXiv:2004.03497}, 2020.

\bibitem[Madani et~al.(2021)Madani, Krause, Greene, Subramanian, Mohr, Holton,
  Olmos, Xiong, Sun, Socher, Fraser, and Naik]{madani2021deep}
Madani, A., Krause, B., Greene, E.~R., Subramanian, S., Mohr, B.~P., Holton,
  J.~M., Olmos, J.~L., Xiong, C., Sun, Z.~Z., Socher, R., Fraser, J.~S., and
  Naik, N.
\newblock Deep neural language modeling enables functional protein generation
  across families.
\newblock \emph{bioRxiv}, 2021.
\newblock \doi{10.1101/2021.07.18.452833}.
\newblock URL
  \url{https://www.biorxiv.org/content/early/2021/07/18/2021.07.18.452833}.

\bibitem[Madani et~al.(2023)Madani, Krause, Greene, Subramanian, Mohr, Holton,
  Olmos~Jr, Xiong, Sun, Socher, et~al.]{madani2023large}
Madani, A., Krause, B., Greene, E.~R., Subramanian, S., Mohr, B.~P., Holton,
  J.~M., Olmos~Jr, J.~L., Xiong, C., Sun, Z.~Z., Socher, R., et~al.
\newblock Large language models generate functional protein sequences across
  diverse families.
\newblock \emph{Nature Biotechnology}, pp.\  1--8, 2023.

\bibitem[Mallinson et~al.(2022)Mallinson, Adamek, Malmi, and
  Severyn]{mallinson2022edit5}
Mallinson, J., Adamek, J., Malmi, E., and Severyn, A.
\newblock {E}di{T}5: Semi-autoregressive text editing with t5 warm-start.
\newblock In \emph{Findings of the Association for Computational Linguistics:
  EMNLP 2022}, pp.\  2126--2138, Abu Dhabi, United Arab Emirates, December
  2022. Association for Computational Linguistics.
\newblock URL \url{https://aclanthology.org/2022.findings-emnlp.156}.

\bibitem[Mueller et~al.(2017)Mueller, Gifford, and
  Jaakkola]{mueller2017sequence}
Mueller, J., Gifford, D., and Jaakkola, T.
\newblock Sequence to better sequence: continuous revision of combinatorial
  structures.
\newblock In \emph{International Conference on Machine Learning}, pp.\
  2536--2544. PMLR, 2017.

\bibitem[Novak et~al.(2016)Novak, Auli, and Grangier]{novak2016iterative}
Novak, R., Auli, M., and Grangier, D.
\newblock Iterative refinement for machine translation.
\newblock \emph{arXiv preprint arXiv:1610.06602}, 2016.

\bibitem[Pang et~al.(2022)Pang, Padmakumar, Sellam, Parikh, and
  He]{pang2022reward}
Pang, R.~Y., Padmakumar, V., Sellam, T., Parikh, A.~P., and He, H.
\newblock Reward gaming in conditional text generation.
\newblock \emph{arXiv preprint arXiv:2211.08714}, 2022.

\bibitem[Raffel et~al.(2022)Raffel, Shazeer, Roberts, Lee, Narang, Matena,
  Zhou, Li, and Liu]{raffel2020exploring}
Raffel, C., Shazeer, N., Roberts, A., Lee, K., Narang, S., Matena, M., Zhou,
  Y., Li, W., and Liu, P.~J.
\newblock Exploring the limits of transfer learning with a unified text-to-text
  transformer.
\newblock \emph{Journal of Machine Learning Research}, 21\penalty0 (1), June
  2022.
\newblock ISSN 1532-4435.

\bibitem[Ren et~al.(2022)Ren, Li, Ding, Zhou, Ma, and Peng]{ren2022proximal}
Ren, Z., Li, J., Ding, F., Zhou, Y., Ma, J., and Peng, J.
\newblock Proximal exploration for model-guided protein sequence design.
\newblock In \emph{International Conference on Machine Learning}, pp.\
  18520--18536. PMLR, 2022.

\bibitem[Romero \& Arnold(2009)Romero and Arnold]{romero2009exploring}
Romero, P.~A. and Arnold, F.~H.
\newblock Exploring protein fitness landscapes by directed evolution.
\newblock \emph{Nature Reviews Molecular Cell Biology}, 10\penalty0
  (12):\penalty0 866--876, 2009.

\bibitem[Russell et~al.(2017)Russell, Bennett, Wellman, Chung, Yu, Tillman,
  Wittes, Pappas, Elci, McCague, et~al.]{russell2017efficacy}
Russell, S., Bennett, J., Wellman, J.~A., Chung, D.~C., Yu, Z.-F., Tillman, A.,
  Wittes, J., Pappas, J., Elci, O., McCague, S., et~al.
\newblock Efficacy and safety of voretigene neparvovec (aav2-hrpe65v2) in
  patients with rpe65-mediated inherited retinal dystrophy: a randomised,
  controlled, open-label, phase 3 trial.
\newblock \emph{The Lancet}, 390\penalty0 (10097):\penalty0 849--860, 2017.

\bibitem[Schymkowitz et~al.(2005)Schymkowitz, Borg, Stricher, Nys, Rousseau,
  and Serrano]{schymkowitz2005foldx}
Schymkowitz, J., Borg, J., Stricher, F., Nys, R., Rousseau, F., and Serrano, L.
\newblock The foldx web server: an online force field.
\newblock \emph{Nucleic Acids Research}, 33\penalty0 (suppl\_2):\penalty0
  W382--W388, 2005.

\bibitem[Shire et~al.(2004)Shire, Shahrokh, and Liu]{shire2004challenges}
Shire, S.~J., Shahrokh, Z., and Liu, J.
\newblock Challenges in the development of high protein concentration
  formulations.
\newblock \emph{Journal of pharmaceutical sciences}, 93\penalty0 (6):\penalty0
  1390--1402, 2004.

\bibitem[Vaswani et~al.(2017)Vaswani, Shazeer, Parmar, Uszkoreit, Jones, Gomez,
  Kaiser, and Polosukhin]{vaswani2017attention}
Vaswani, A., Shazeer, N., Parmar, N., Uszkoreit, J., Jones, L., Gomez, A.~N.,
  Kaiser, {\L}., and Polosukhin, I.
\newblock Attention is all you need.
\newblock \emph{Advances in Neural Information Processing Systems}, 30, 2017.

\bibitem[Verkuil et~al.(2022)Verkuil, Kabeli, Du, Wicky, Milles, Dauparas,
  Baker, Ovchinnikov, Sercu, and Rives]{verkuil2022language}
Verkuil, R., Kabeli, O., Du, Y., Wicky, B.~I., Milles, L.~F., Dauparas, J.,
  Baker, D., Ovchinnikov, S., Sercu, T., and Rives, A.
\newblock Language models generalize beyond natural proteins.
\newblock \emph{bioRxiv}, 2022.

\bibitem[Wang(1999)]{wang1999instability}
Wang, W.
\newblock Instability, stabilization, and formulation of liquid protein
  pharmaceuticals.
\newblock \emph{International journal of pharmaceutics}, 185\penalty0
  (2):\penalty0 129--188, 1999.

\bibitem[Webber et~al.(2016)Webber, Appel, Vinciguerra, Cortinas, Thapa,
  Jhunjhunwala, Isaacs, Langer, and Anderson]{webber2016supramolecular}
Webber, M.~J., Appel, E.~A., Vinciguerra, B., Cortinas, A.~B., Thapa, L.~S.,
  Jhunjhunwala, S., Isaacs, L., Langer, R., and Anderson, D.~G.
\newblock Supramolecular pegylation of biopharmaceuticals.
\newblock \emph{Proceedings of the National Academy of Sciences}, 113\penalty0
  (50):\penalty0 14189--14194, 2016.

\bibitem[Welleck et~al.(2023)Welleck, Lu, West, Brahman, Shen, Khashabi, and
  Choi]{welleck2022generating}
Welleck, S., Lu, X., West, P., Brahman, F., Shen, T., Khashabi, D., and Choi,
  Y.
\newblock Generating sequences by learning to self-correct.
\newblock In \emph{The Eleventh International Conference on Learning
  Representations}, 2023.
\newblock URL \url{https://openreview.net/forum?id=hH36JeQZDaO}.

\bibitem[Wolf et~al.(2020)Wolf, Debut, Sanh, Chaumond, Delangue, Moi, Cistac,
  Rault, Louf, Funtowicz, Davison, Shleifer, von Platen, Ma, Jernite, Plu, Xu,
  Le~Scao, Gugger, Drame, Lhoest, and Rush]{wolf-etal-2020-transformers}
Wolf, T., Debut, L., Sanh, V., Chaumond, J., Delangue, C., Moi, A., Cistac, P.,
  Rault, T., Louf, R., Funtowicz, M., Davison, J., Shleifer, S., von Platen,
  P., Ma, C., Jernite, Y., Plu, J., Xu, C., Le~Scao, T., Gugger, S., Drame, M.,
  Lhoest, Q., and Rush, A.
\newblock Transformers: State-of-the-art natural language processing.
\newblock In \emph{Proceedings of the 2020 Conference on Empirical Methods in
  Natural Language Processing: System Demonstrations}, pp.\  38--45, Online,
  October 2020. Association for Computational Linguistics.
\newblock \doi{10.18653/v1/2020.emnlp-demos.6}.
\newblock URL \url{https://aclanthology.org/2020.emnlp-demos.6}.

\bibitem[Yang \& Klein(2021)Yang and Klein]{yang-klein-2021-fudge}
Yang, K. and Klein, D.
\newblock {FUDGE}: Controlled text generation with future discriminators.
\newblock In \emph{Proceedings of the 2021 Conference of the North American
  Chapter of the Association for Computational Linguistics: Human Language
  Technologies}, pp.\  3511--3535, Online, June 2021. Association for
  Computational Linguistics.
\newblock \doi{10.18653/v1/2021.naacl-main.276}.
\newblock URL \url{https://aclanthology.org/2021.naacl-main.276}.

\bibitem[Yang et~al.(2019)Yang, Wu, and Arnold]{yang2019machine}
Yang, K.~K., Wu, Z., and Arnold, F.~H.
\newblock Machine-learning-guided directed evolution for protein engineering.
\newblock \emph{Nature Methods}, 16\penalty0 (8):\penalty0 687--694, 2019.

\bibitem[Zhang et~al.(2015)Zhang, Zhao, and LeCun]{zhang2015character}
Zhang, X., Zhao, J., and LeCun, Y.
\newblock Character-level convolutional networks for text classification.
\newblock \emph{Advances in Neural Information Processing Systems}, 28, 2015.

\end{thebibliography}
\bibliographystyle{icml2023}

\newpage
\appendix
\onecolumn
\section{Limitations}
\label{sec:limitations}

\paragraph{Creation of synthetic data can introduce hallucinations in natural language} Our method relies on masked language modeling to create minimally perturbed pairs of sequences (\Cref{sec:train_ice}). In natural language tasks, this can result in a perturbed sequence that is slightly different in meaning from the source sequence. As a result, the \modelacc model when trained can also alter the meaning of the sequence. In particular, we want to note that certain kinds of hallucinations from text generation models can be harmful if used without proper consideration. Specifically, in \Cref{tab:example_sentiment}, it is acceptable for the model to edit the sentiment associated with the food or ambiance at the restaurant but we want the model to retain the basic information that the writer and his partner are eating at a sushi restaurant in Scottsdale.
Going forward, we intend to investigate better strategies for synthetic data creation to measure and mitigate this occurrence.

\paragraph{Assumption that edits in the \emph{training region} generalize to \emph{extrapolation region}} Our work relies on training a model on perturbations made on sequences belonging to the \emph{training region}. We then repeatedly make edits to increase or decrease the score into the \emph{extrapolation region}. While our experiments show promising results, we believe that this assumption does not equally hold for all tasks and domains. We intend to study this further going forward.  

\paragraph{Relying on trained models to score sequences} For evaluation of the sentiment control and the AAV tasks, we train classifier models to measure the attribute values of the sequences. These models only estimate the ground truth attribute values and can end up learning spurious correlations from the datasets. We note that these are to be used as a means to benchmark our method against the various baselines. Particularly in the case of proteins such as AAV, prior to any real-world usage, a detailed analysis of the oracle models or real-life wet lab experiments should be performed.

\paragraph{Inference for iterative methods is slow} By the nature of our method, iteratively editing a sequence is much slower in terms of inference time as compared to a single-step edit by a model such as \emph{Genhance}.

\section{Additional Model Training Details}
\label{sec:expt_details}
We fine-tune all of the language models for our experiments using the HuggingFace library \cite{wolf-etal-2020-transformers}. All of the code used for our experiments and trained models is available at \url{https://github.com/vishakhpk/iter-extrapolation}.

\paragraph{Sentiment Control} The scorer and oracle model used for evaluation are fine-tuned RoBERTa-Large \cite{liu2019roberta} models. The oracle is trained on the entire Yelp dataset. The scorer is trained on those examples with a sentiment from $2$ to $4$. Both the scorer and oracle are fine-tuned to optimize the mean-squared error loss on the gold labels from the dataset.
We create paired data to train the \modelacc generator model using the scorer and a pre-trained T5-Base \cite{raffel2020exploring} model. We create $100$K pairs and fine-tune T5-Base to serve as the \modelacc generator. The hyperparameter $\delta=0.4$ used to filter synthetic pairs was selected based on a small internal pilot. We fine-tune T5-Base to generate the output of the synthetic pairs given the input sequences optimizing the cross-entropy loss on the output tokens. 
For each of these, we use the recommended hyperparameters from the HuggingFace repository and sweep learning rates from $1e{-6}$ to $1e{-3}$. 

\paragraph{ACE2} For ACE2, we fine-tune a {ProtBert} \cite{elnaggar2020prottrans} model, made available via the HuggingFace, to predict the \emph{ddG} values given the mutants from the dataset released by \citet{chan2021deep}. Here we optimize the mean-squared error loss on the gold labels, selecting the optimum checkpoint using the validation loss. We use this to create a synthetic dataset of $1$M pairs which is used to fine-tune the \modelacc generator model. We fine-tune Prot-T5-XL \cite{elnaggar2020prottrans} on these pairs to generate the output of the synthetic pairs given the input sequences optimizing the cross-entropy loss on the output tokens. We again use the recommended hyperparameters from the HuggingFace repository and sweep learning rates from $1e{-6}$ to $1e{-3}$. For scoring with FoldX, we match the parameters from \cite{chan2021deep}.

\paragraph{AAV} The scorer and oracle models for the AAV task are CNN models that accept the protein sequence as a string and output a real number corresponding to the fitness value. We select the model architecture according to the parameters specified in the FLIP benchmark \cite{dallago2022flip}. We follow the same as the obtained the highest test spearman correlation for the AAV \emph{low-vs-high} split. Both CNN models are trained from the repository of the benchmark optimizing the mean squared-error loss on the fitness values. We use the scorer to create $1$M synthetic pairs to train the \modelacc generator model optimizing the cross-entropy loss of the output tokens given the input protein sequence and corresponding control tag. 

\section{Additional Findings}
\label{app:add_results}

\subsection{Exploring Diversity in AAV Mutants}
\label{app:aav_results_extra}
\begin{figure*}[ht!]
     \centering
     \begin{subfigure}[b]{0.475\textwidth}
   \centering
   \includegraphics[width=\textwidth]{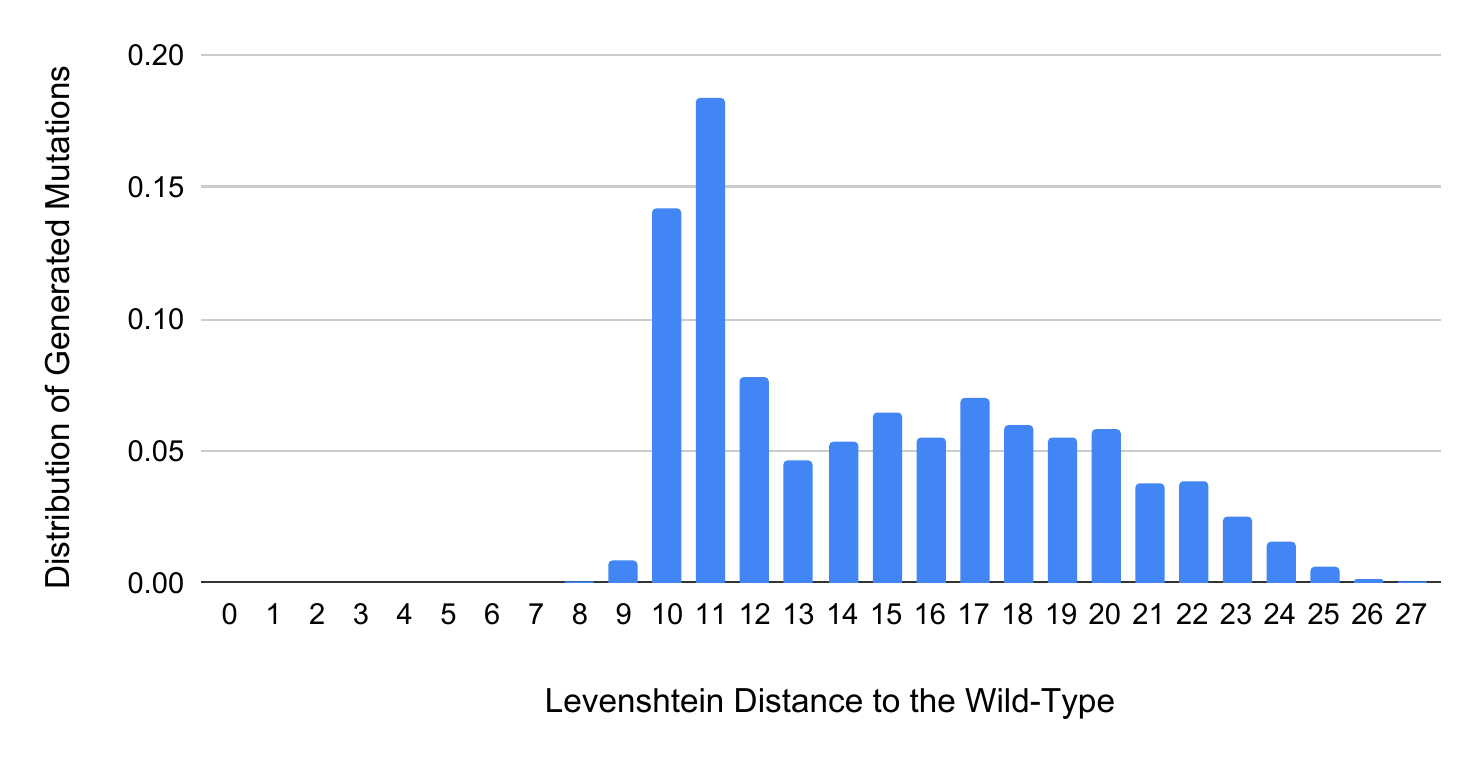}
   \caption{}
   \label{fig:aav-dist-w-disc}
     \end{subfigure}
     \hfill
     \begin{subfigure}[b]{0.475\textwidth}
   \centering
   \includegraphics[width=\textwidth]{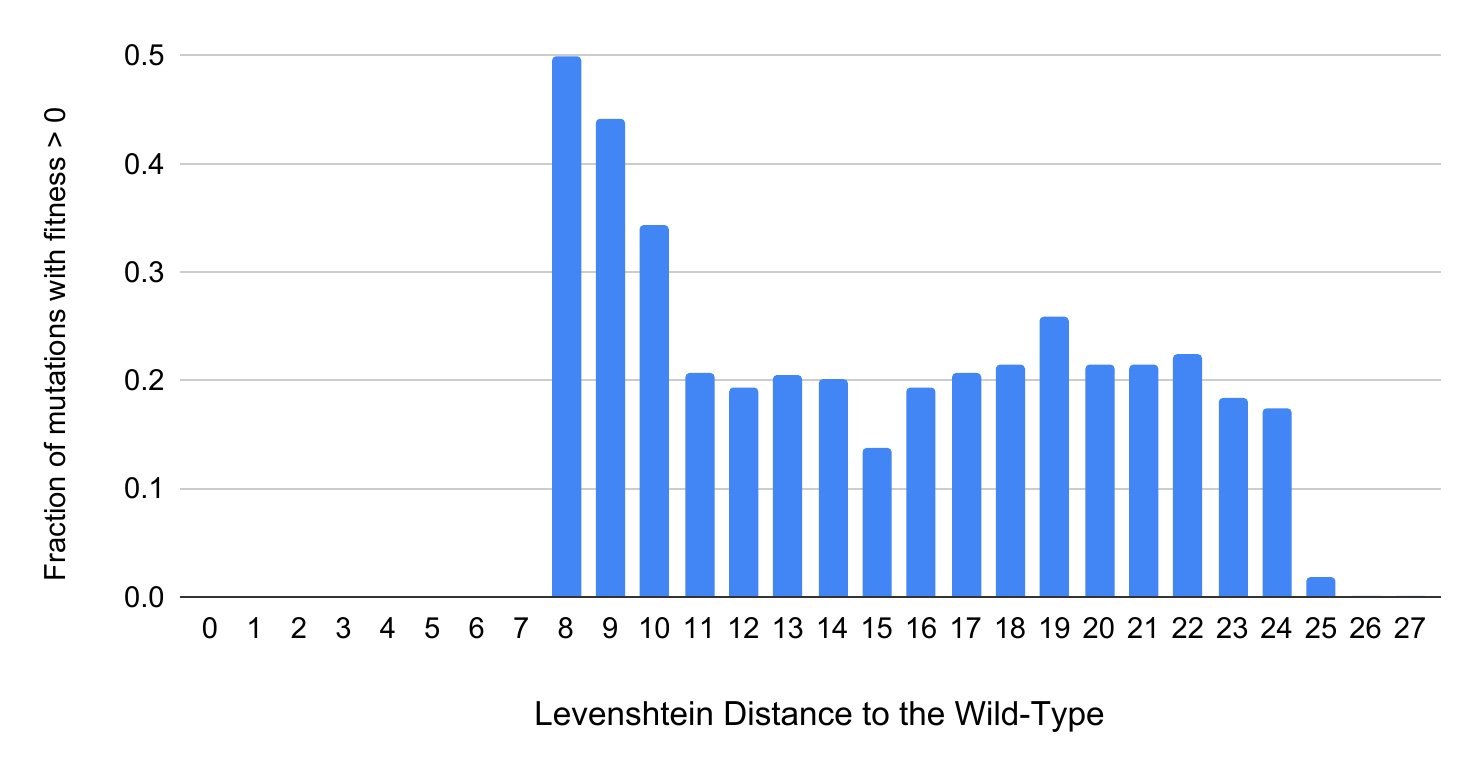}
   \caption{}
   \label{fig:aav-dist-w-success-rates}
     \end{subfigure}
     \caption{We plot the fraction of sequences for a given Levenshtein distance away from the wild type (\Cref{fig:aav-dist-w-disc}). \Cref{fig:aav-dist-w-success-rates} shows the fraction of generated sequences that are better than the wild type (according to the oracle) as a function of the Levenshtein distance, showing the potential of \modelacc to generate both diverse and viable mutants.}
     \label{fig:output_dist_aav_app}
\end{figure*}

While AAV capsids hold promise for gene therapy, the immunity from prior AAV exposure excludes 20--80\% of the population from such treatments~\cite{calcedo2009worldwide}. Thus, it is essential to not only generate AAV mutants of high fitness, but also of significant diversity from the wild type. To this end, in Figure~\ref{fig:output_dist_aav}, we analyze the distribution of sequences generated by our model (in the $10$th iteration) as a function of their Levenshtein distance from the wild-type. 
We see that while the majority of mutations generated have an edit distance of around $8$--$10$, the model generates mutations having as far as $25$ edits from the wild-type (\Cref{fig:aav-dist-w-disc}). However, we see that even when the model makes over $20$ edits, the fraction of examples within this bucket is still $0.2$, showing a large diversity in the mutations generated (\Cref{fig:aav-dist-w-success-rates}).
 
We note that the model generates a mutant at a diverse range of levenstein distances from the wild type (8 to 27). Moreover, \modelacc displays strong performance throughout this range according to our oracle (Figure~\ref{fig:aav-dist-w-success-rates}), demonstrating its potential to generate both viable and diverse mutants of AAV.

\subsection{Additional Results on Sentiment Control}
\label{app:sent_results_extra}
\Cref{tab:example_sentiment} shows an example of the editing process, increasing the sentiment score of the input review iteratively. 
In addition to the results from \Cref{tab:sent_results}, we report a few variants of \modelacc and \emph{Genhance}. For \modelacc, the masking strategy to create synthetic paired data involves sampling a location in the sequence to start the mask using a Bernoulli distribution ($p=0.8$) and then selecting the length of the mask (in terms of tokens masked) by sampling from a truncated Poisson distribution. The results presented in \Cref{tab:sent_results} correspond to the \emph{Super Large} variant in \Cref{tab:sent_results_full} where $\lambda = 6$ and the maximum span size is set to $12$. We also report three other variants of the masking strategy \emph{Small} ($\lambda = 3$, maximum of 6), \emph{Medium} ($\lambda = 4$, maximum of 8) and \emph{Large} ($\lambda = 5$, maximum of 10). We observed the best extrapolation results on the \emph{Super Large} variant and used this masking strategy to report the \emph{Sampling} and \emph{Iterative Sampling} baselines. We also report two variants of \emph{Genhance} where we vary the total number of output sequences generated for each example. As we increase $n$, the model predictably performs better at extrapolation but we see that the directly comparable variant, $n=50$, is outperformed by \modelacc.

\begin{table*}[ht!]
\centering
\begin{tabular}{R{5cm}|ccc|ccc}
\toprule
\multirow{2}{*}{\textbf{Target Sentiment Score}} & \multicolumn{3}{c|}{\textbf{Training Region}}  & \multicolumn{3}{c|}{\textbf{Extrapolation Region}} \\ 
 & \multicolumn{1}{c}{\textbf{3.5}} & \multicolumn{1}{c}{\textbf{2.5}} & \textbf{Average} & \multicolumn{1}{c}{\textbf{4.5}} & \multicolumn{1}{c}{\textbf{1.5}} & \textbf{Average} \\ \midrule
Score-Conditioned Baseline   & 0.780   & 0.766   & 0.773    & 0.212  & 0.217   & 0.215        \\
PPLM & 0.534 & 0.516 & 0.522 & 0.081 & 0.065 & 0.077 \\
Sampling & 0.362 & 0.259 & 0.310 & 0.061 & 0.050 & 0.056 \\
Iterative Sampling & 0.668 & 0.657 & 0.663 & 0.320 & 0.328 & 0.324 \\
\midrule
Genhance (n = 1)   & 0.407   & 0.167   & 0.287    & 0.063         & 0.025   & 0.044        \\
Genhance (n = 50)  & 0.982   & 0.833   & 0.908 & 0.482   & 0.291   & 0.387         \\
Genhance (n = 100)         & \textbf{0.995}   & 0.912   & \textbf{0.954}         & \textbf{0.670}         & 0.429   & 0.550        \\
\midrule
\modelacc w/ Scorer -- Small    & 0.962   & 0.98    & 0.971    & 0.514   & 0.344   & 0.429   \\
Medium & 0.945   & 0.870   & 0.908          & 0.636   & 0.499   & 0.567   \\
Large & 0.953 & 0.884 & 0.918 & 0.649 & 0.555 & 0.602 \\
Super Large   & 0.943   & 0.900   & 0.921    & 0.638   & \textbf{0.582}   & \textbf{0.610} \\
\midrule
\modelacc Scorer-Free & 0.976   & \textbf{0.918}   & 0.947    & 0.446   & 0.305   & 0.376 \\
\bottomrule
\end{tabular}
\caption{Results on sentiment control in both the training and extrapolation regions including ablations of our model and Genhance. Evaluation is done by measuring the fraction of examples that have a sentiment value greater than (or less than) a target score as determined by the oracle scorer. Bold values are the highest success rates for each target. \modelacc achieves the highest rate of extrapolation.}
\label{tab:sent_results_full}
\end{table*}

\begin{table}[]
\centering
\begin{tabular}{c|C{15cm}}
\toprule
Iteration & Text\\ \midrule
1         & \textless{}inc\textgreater This place is just okay. The wait is way too long, the counter person is irritating, and the unsupervised children are disruptive. The fill and nail art are decent, but not worth the wait. Lisa was fast, and I dried quickly. I found the nail art I wanted on Pinterest, and she did a decent job replicating it. She also charged my phone, which was thoughtful. If you're going, try to get there early, as it gets crowded later in the day. It's not amazing, but it's worth checking out. \\ \midrule
2         & \textless{}inc\textgreater This place is average at best. The wait is too long, the counter person is annoying, and the unsupervised children are a problem. The fill and nail art are decent, but nothing special. Lisa was quick, and I dried quickly. I found the nail art I wanted on Pinterest, and she did a good job replicating it. She also charged my phone, which was nice. If you're going, try to get there early, as it gets busier as the day goes on. It's worth a visit, but don't expect anything amazing.
\\ \midrule
3         & \textless{}inc\textgreater This place is decent, but has some flaws. The wait is long, the counter person is too pushy, and the unsupervised children are distracting. The fill and nail art are really good, though, and Lisa was super quick. I found the nail art I wanted on Pinterest, and she replicated it perfectly. Bonus points for charging my phone. If you're thinking about going, try to get there early, as the line gets longer later in the day. Despite the flaws, it's still worth a visit. \\ \midrule
4         & \textless{}inc\textgreater This place is good, but not great. The wait is a bit long, the counter person is pushy, and the unsupervised children are annoying. However, the fill and nail art are fantastic, so it's worth it. Lisa was quick, and I was drying in no time. I found the nail art I wanted on Pinterest, and she copied it perfectly. She even charged my phone, which was nice. If you plan to visit, go early as it gets busier closer to closing time.
\\ \midrule
5         & \textless{}inc\textgreater This place is pretty good! The wait, pushy counter person, and unsupervised children are easy to ignore when I leave with the best fill and nail art ever! Lisa was lightning-fast, and I was drying in under 15 minutes! I found the nail art I wanted on Pinterest, and she copied it flawlessly! As a bonus, she charged my phone, which was a nice gesture! If you're planning on going, try to arrive early, as the line gets longer closer to closing. But trust me, it's worth the wait! \\ \midrule
6         & \textless{}inc\textgreater This place is great! The wait, pushy counter person, and unsupervised children are an easy overlook when I finally leave with the best fill and nail art I've ever had! Lisa was super quick, had me drying in less than 15 minutes of sitting down in her chair! I found the nail art I wanted (she copied it perfectly, by the way) on pintrest, but just as I sat down, my phone died. She pulled out her charger, and charged my phone! Where else has anyone done this? Nowhere. Just a heads up, go early, if you can, as it gets closer to close, more and more people line up. :) it's so worth the wait, though!! \\ \bottomrule
\end{tabular}
\caption{Trajectory of improving the sentiment associated with a review using \modelacc.}
\label{tab:example_sentiment}
\end{table}

\subsection{Sensitivity to Hyperparameters of Generation}
\label{app:hyperparameter_experiment}

To study the interaction between the generation hyperparameters and the number of iterations at inference time, we ran both scorer-free inference varying the beam size and scorer-guided inference varying $k$ in top-$k$ for the \emph{ACE2} task. In all cases, we generated $1000$ mutations. We present the results at iteration $2$,$5$, and $10$ in \Cref{tab:ace2_hyperparam}. Each cell of the table represents the fraction of mutations with \emph{ddG} value lower than the corresponding target rounded off to three decimal places. The rows corresponding to top-$k = 5$ and beam size $5$ at iteration $10$ were included in \Cref{tab:ace2_results}.

Overall, we find that the results at the end of the inference process (iteration $10$) are largely stable w.r.t. these hyperparameters. In particular, when increasing $k$ for top-$k$ sampling, we see a slight drop in performance, which might be due to the small vocabulary size of protein sequences (a total of $20$). Similarly, for scorer-free inference, as we decrease beam size to $3$ we obtain slightly better performance in the training region with a small drop-off for extrapolation. Increasing the beam size to $10$ mildly decreases performance. 

We find that the iteration number is a reliable indicator of the extrapolation performance with little change in performance observed due to the top-$k$ and beam size hyperparameters (within each specific iteration). At iteration 2, when guided by the scorer, a higher top-$k$ value results in better performance as the model samples more diverse generations, and the scorer can reliably select good sequences to obtain better performance on targets in the training region. Similarly, for scorer-free inference, a higher beam size also improves performance on the targets in the training region. However as we increase the number of iterations to iteration $5$ and $10$, this effect largely evens out. 

\begin{table}[ht!]
\small
\centering
\begin{tabular}{crr|cc|ccc}
\toprule
\multicolumn{3}{c|}{\multirow{2}{*}{\textbf{Target \emph{ddG} Value}}} & \multicolumn{2}{c|}{\textbf{Training Region}} & \multicolumn{3}{c}{\textbf{Extrapolation Region}} \\
\multicolumn{3}{c|}{} & \textbf{-1} & \textbf{-2.5} & \textbf{-5} & \textbf{-6} & \textbf{-7} \\ \midrule
\multicolumn{1}{c|}{\multirow{9}{*}{\textbf{ICE w/ Scorer: Varying K for sampling}}} & \multicolumn{1}{r|}{\multirow{3}{*}{\textbf{Iteration = 10}}} & \textbf{TopK = 15} & 0.997 & 0.964 & 0.249 & 0.083 & 0.01 \\
\multicolumn{1}{c|}{} & \multicolumn{1}{r|}{} & \textbf{TopK = 10} & 0.998 & 0.966 & 0.283 & 0.091 & 0.016 \\
\multicolumn{1}{c|}{} & \multicolumn{1}{r|}{} & \textbf{TopK = 5} & \textbf{0.998} & \textbf{0.974} & \textbf{0.362} & \textbf{0.098} & \textbf{0.019} \\ \cmidrule{2-8} 
\multicolumn{1}{c|}{} & \multicolumn{1}{r|}{\multirow{3}{*}{\textbf{Iteration = 5}}} & \textbf{TopK = 15} & 0.982 & 0.648 & 0.041 & 0.004 & 0.000 \\
\multicolumn{1}{c|}{} & \multicolumn{1}{r|}{} & \textbf{TopK = 10} & 0.981 & 0.646 & 0.040 & 0.004 & 0.000 \\
\multicolumn{1}{c|}{} & \multicolumn{1}{r|}{} & \textbf{TopK = 5} & 0.978 & 0.647 & 0.042 & 0.005 & 0.001 \\ \cmidrule{2-8} 
\multicolumn{1}{c|}{} & \multicolumn{1}{r|}{\multirow{3}{*}{\textbf{Iteration = 2}}} & \textbf{TopK = 15} & 0.711 & 0.093 & 0.002 & 0.000 & 0.000 \\
\multicolumn{1}{c|}{} & \multicolumn{1}{r|}{} & \textbf{TopK = 10} & 0.703 & 0.090 & 0.001 & 0.000 & 0.000 \\
\multicolumn{1}{c|}{} & \multicolumn{1}{r|}{} & \textbf{TopK = 5} & 0.674 & 0.086 & 0.001 & 0.000 & 0.000 \\ \midrule
\multicolumn{1}{c|}{\multirow{9}{*}{\textbf{ICE Scorer-Free: Varying beam size}}} & \multicolumn{1}{r|}{\multirow{3}{*}{\textbf{Iteration = 10}}} & \textbf{Beam Size = 10} & 0.930 & 0.572 & 0.059 & 0.013 & 0.000 \\
\multicolumn{1}{c|}{} & \multicolumn{1}{r|}{} & \textbf{Beam Size = 5} & 0.945 & 0.598 & \textbf{0.062} & \textbf{0.017} & \textbf{0.002} \\
\multicolumn{1}{c|}{} & \multicolumn{1}{r|}{} & \textbf{Beam Size = 3} & \textbf{0.959} & \textbf{0.623} & 0.060 & 0.016 & 0.000 \\ \cmidrule{2-8} 
\multicolumn{1}{c|}{} & \multicolumn{1}{r|}{\multirow{3}{*}{\textbf{Iteration = 5}}} & \textbf{Beam Size = 10} & 0.852 & 0.440 & 0.030 & 0.006 & 0.000 \\
\multicolumn{1}{c|}{} & \multicolumn{1}{r|}{} & \textbf{Beam Size = 5} & 0.847 & 0.437 & 0.026 & 0.005 & 0.000 \\
\multicolumn{1}{c|}{} & \multicolumn{1}{r|}{} & \textbf{Beam Size = 3} & 0.844 & 0.419 & 0.023 & 0.004 & 0.000 \\ \cmidrule{2-8} 
\multicolumn{1}{c|}{} & \multicolumn{1}{r|}{\multirow{3}{*}{\textbf{Iteration = 2}}} & \textbf{Beam Size = 10} & 0.620 & 0.182 & 0.001 & 0.000 & 0.000 \\
\multicolumn{1}{c|}{} & \multicolumn{1}{r|}{} & \textbf{Beam Size = 5} & 0.567 & 0.155 & 0.001 & 0.000 & 0.000 \\
\multicolumn{1}{c|}{} & \multicolumn{1}{r|}{} & \textbf{Beam Size = 3} & 0.526 & 0.143 & 0.000 & 0.000 & 0.000 \\ \bottomrule
\end{tabular}
\caption{Evaluation on the \emph{ACE2} task to study the interaction between the generation hyperparameters and the number of iterations at inference time. Each table cell represents the fraction of mutations with a \emph{ddG} value lower than the corresponding target. We vary $k$ for top-$k$ sampling for scorer-guided inference and vary beam size for scorer-free inference. We find that the results are largely stable with respect to these hyperparameters at the end of inference (\ie iteration $10$). Early on during inference (\ie iteration $2$), we find that a higher top-$k$ value and beam size respectively result in better performance but this largely evens out by iteration $5$ and $10$.} 
\label{tab:ace2_hyperparam}
\end{table}

\subsection{Stopping Criteria}
\label{sec:stopping}
Reliably identifying when the generation model has reached a target score is difficult due to the extrapolative nature of the task. Specifically, if we had a way to know when the generator model has achieved a target score in the extrapolation region, then this supervision could directly be used to train the generator itself. One option is to use the scorer, $f_s$. However, we observed the output of $f_s$ plateau near the boundary of the training region, limiting its reliability as a stopping condition in the extrapolation region. To illustrate this, we present the average output score in the \emph{ACE2} task as a function of $10$ iterations  in \Cref{tab:stopping}. We observed that the output score remained largely constant beyond iteration $7$. Hence we settled on setting the number of iterations to greater than the plateau point of the scorer, such as $10$, and found that this worked well across our $3$ datasets without further tuning. However, we acknowledge the need for a more principled stopping condition as an open problem in this setting.

\begin{table}[ht!]
\centering
    \begin{tabular}{|c|c|}
    \hline
    \textbf{Iteration} & \textbf{Average Score} \\ \hline
    1 & -0.673 \\ \hline
    2 & -2.051 \\ \hline
    3 & -2.879 \\ \hline
    4 & -3.272 \\ \hline
    5 & -3.446 \\ \hline
    6 & -3.522 \\ \hline
    7 & -3.551 \\ \hline
    8 & -3.558 \\ \hline
    9 & -3.555 \\ \hline
    10 & -3.567 \\ \hline
    \end{tabular}
\caption{Average output scores of $f_s$ as a function of iterations in the \emph{ACE2} task. Each cell is an average of the scores assigned to the $10,000$ mutants generated with scorer-guided inference in \Cref{tab:ace2_results}. We observe that the output of $f_s$ plateaus near the boundary of the training region at around $-3.5$ making it unreliable as a stopping condition for the generation process.}
\label{tab:stopping}
\end{table}


\end{document}